\documentclass[lettersize,journal]{IEEEtran}
\usepackage{amsmath,amsfonts}
\usepackage{array}
\usepackage[caption=false,font=normalsize,labelfont=sf,textfont=sf]{subfig}
\usepackage{textcomp}
\usepackage{stfloats}
\usepackage{url}
\usepackage{verbatim}
\usepackage{graphicx}
\usepackage{float}
\usepackage{booktabs}
\usepackage{multirow}
\usepackage{array}
\usepackage{amsmath}
\usepackage{mathtools}
\usepackage{algorithm}
\usepackage{algpseudocode}
\usepackage{amssymb,amsmath,amsthm}

\usepackage{comment}
\usepackage[table]{xcolor}
\usepackage{colortbl}
\usepackage[hidelinks]{hyperref}
\usepackage{orcidlink}
\usepackage{pifont}%%for \ding{55}, namely crossmark
\usepackage{makecell} %%for multiple lines in a cell ybder tabular environ.

\definecolor{OverallAugGenericBg}{RGB}{226,246,238}
\definecolor{OverallAugHealthBg}{RGB}{212,239,228}
\definecolor{OverallAugFinanceBg}{RGB}{198,231,216}
\definecolor{OverallAugRelBg}{RGB}{184,223,204}
\definecolor{OverallImpBg}{RGB}{224,236,255}
\definecolor{OverallTrustBg}{RGB}{255,235,205}
\definecolor{OverallAnomBg}{RGB}{239,229,252}
\DeclareRobustCommand{\taskshade}[2]{\begingroup\setlength{\fboxsep}{1pt}\colorbox{#1}{\strut #2}\endgroup}
\newcommand{\codeurl}[1]{\href{#1}{\raisebox{-0.15ex}{\includegraphics[draft=false,height=1.05em]{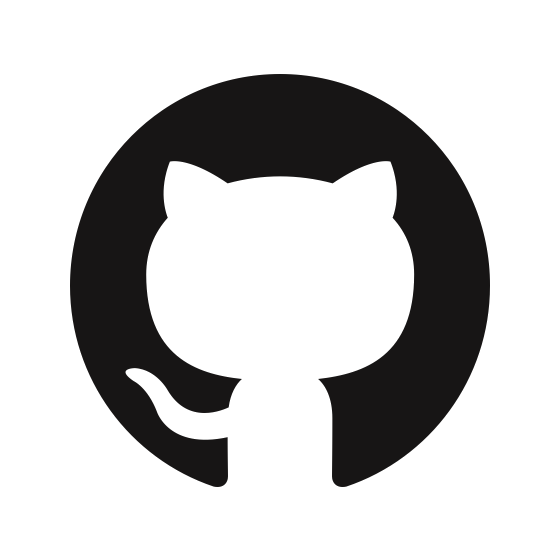}}}}

\newtheorem{problem}{Problem}

\def\BibTeX{{\rm B\kern-.05em{\sc i\kern-.025em b}\kern-.08em
    T\kern-.1667em\lower.7ex\hbox{E}\kern-.125emX}}
\usepackage{balance}
\begin{document}
\title{Diffusion and Flow Matching Models \\for Tabular Data: A Survey}

%\author{Anonymous authors}

%\begin{comment}
% \author{Zhong Li \orcidlink{0000-0003-1124-5778}, Qi Huang* \orcidlink{0009-0007-4989-135X}, Lincen Yang* \orcidlink{0000-0003-1936-2784} \thanks{* Qi Huang and Lincen Yang contributed equally.}, Jiayang Shi \orcidlink{0000-0002-7014-0805}, Zhao Yang  -- I personally think the equal contributions comment for 2nd and 3rd author does not make any difference for anyone; IMHO this is only useful for indicating multiple first authors. - Matthijs
\author{Zhong Li \orcidlink{0000-0003-1124-5778}, Qi Huang \orcidlink{0009-0007-4989-135X}, Lincen Yang* \orcidlink{0000-0003-1936-2784}, Jiayang Shi 
	\orcidlink{0000-0002-7014-0805}, Zhao Yang 
\orcidlink{0000-0002-7011-4340},\\ Niki van Stein \orcidlink{0000-0002-0013-7969} (IEEE Senior Member), Thomas Bäck \orcidlink{0000-0001-6768-1478} (IEEE Fellow), Matthijs van Leeuwen \orcidlink{0000-0002-0510-3549}
\thanks{*Lincen Yang is the corresponding author (l.yang@liacs.leidenuniv.nl). Zhong Li is with Great Bay University and Zhao Yang is with Vrije Universiteit Amsterdam. All other authors are with LIACS, Leiden University, the Netherlands.}}
%\end{comment}

\markboth{Submitted to IEEE for possible publication}%
{How to Use the IEEEtran \LaTeX \ Templates}

\maketitle

\begin{abstract}
Deep generative models have made rapid progress in image, text, audio, and video generation, and are increasingly being applied to structured records. For tabular data, however, generative modeling remains difficult: a dataset may contain numerical and categorical attributes, missing values, sensitive fields, imbalanced categories, complex feature dependencies, and domain constraints. Earlier tabular data modeling methods based on GANs or VAEs have achieved useful results, but they can suffer from unstable training, mode collapse, weak modeling of multimodal distributions, and fragile handling of mixed-type features. Diffusion models have therefore attracted growing interest because their noising-and-denoising formulation provides a flexible and stable way to model complex data distributions, and has been adapted to tabular synthesis, missing-value imputation, trustworthy data generation, and anomaly detection. Flow matching offers a closely related route by learning transport vector fields along probability paths, often with more direct control over path design and sampling efficiency. Despite this progress, the literature on diffusion and flow matching models for tabular data remains difficult to compare because methods target different tasks and rely on different representations, objectives, evaluation protocols, and domain assumptions. To the best of our knowledge, this is the first survey dedicated specifically to diffusion and flow matching models for tabular data. We review work from June 2015 to May 2026, organize it around data-engineering challenges, tasks, design choices, and evaluation dimensions, and discuss open problems in scalability, feature dependency modeling, privacy, fairness, benchmarking, and constraint-aware generation. We maintain updates in a \href{https://github.com/Diffusion-Model-Leiden/awesome-diffusion-models-for-tabular-data}{GitHub repository}.

%We can take \cite{croitoru2023diffusion} as an example.

\end{abstract}


 \section{Introduction}
 \label{sec:introdcution}

Tabular data is a data modality in which information is organized into rows, representing individual records, and columns, representing features or attributes. It is ubiquitous in real-world domains such as healthcare~\cite{yoo2012data}, finance~\cite{dixon2020machine}, education~\cite{algarni2016data}, transportation~\cite{anand2018extensive}, and psychology~\cite{king2014data}. The demand for high-quality generative models in these domains is acute due to data privacy regulations such as GDPR~\cite{gdpr2016general} and CCPA~\cite{california2018consumer}. Consequently, real user data is often restricted from public release, whereas synthetic data generated by generative models can preserve machine learning utility while being more easily shared~\cite{kotelnikov2023tabddpm}. Beyond privacy concerns, real-world tabular datasets often contain missing values caused by human errors, survey non-response, data integration failures, or sensor malfunction. Generative models have therefore been employed for missing-value imputation~\cite{mattei2019miwae}. In addition, class imbalance is common in tabular applications~\cite{kaur2019systematic}, and synthetic samples can be used to augment minority classes~\cite{fajardo2021oversampling}. These applications underscore the growing importance of generative modeling for tabular data, ranging from privacy protection~\cite{assefa2020generating,hernandez2022synthetic}, missing-value imputation~\cite{you2020handling,yoon2018gain}, to training data augmentation~\cite{fonseca2023tabular}.

Deep generative models for tabular data include Energy-based Models (EBMs)~\cite{lecun2006tutorial}, Variational Autoencoders (VAEs)~\cite{kingma2013auto}, Generative Adversarial Networks (GANs)~\cite{goodfellow2014generative}, Autoregressive Models~\cite{vaswani2023attentionneed}, Normalizing Flows~\cite{kobyzev2020normalizing}, Diffusion Models~\cite{sohl2015deep}, and Flow Matching models~\cite{lipman2023flow}. Diffusion models have become especially influential because they avoid several common difficulties of earlier generative models: they are generally more stable to train than GANs, less prone to mode collapse, and better suited to representing complex multimodal distributions than simple latent-variable formulations. Flow matching shares the continuous-time generative modeling viewpoint, but learns transport vector fields along probability paths rather than reversing a fixed noising process. Together, diffusion and flow matching provide flexible tools for modeling complex data distributions, while leaving important design choices about representation, probability paths, sampling, privacy risk, and evaluation.

Motivated by their success in images~\cite{ho2020denoising,songscore}, audio~\cite{chen2020wavegrad,kong2020diffwave}, text~\cite{hoogeboom2021argmax,austin2021structured}, video~\cite{xing2024survey}, and graphs~\cite{liu2023generative}, recent studies have begun adapting diffusion and flow matching models to tabular data~\cite{kimstasy,kotelnikov2023tabddpm,suh2023autodiff,lee2023codi,zhangmixed,jolicoeur2024generating,guzman-cordero2025exponential,nasution2026flow,anonymous2026tabflowm,mueller2026cascaded,branco2026patientflow}. These works suggest clear potential, but tabular data also poses distinctive challenges: missing values, non-Gaussian and multimodal numerical distributions, heterogeneous and imbalanced categorical features, mixed-type columns, feature dependencies, limited data availability, privacy constraints, and domain-specific validity rules. As a result, models designed for images, text, or audio cannot be directly transferred to tabular data without careful changes in representation, objective design, conditioning, sampling, post-processing, and evaluation.

Numerous surveys have summarized diffusion models across domains~\cite{yang2023diffusion,cao2024survey} or for specific modalities such as images~\cite{croitoru2023diffusion}, text~\cite{zhu2023diffusion}, videos~\cite{xing2024survey}, time series~\cite{lin2024diffusion}, and graphs~\cite{liu2023generative}. Surveys on tabular generative models~\cite{wang2024challenges,kim2024generative,shi2025comprehensive} provide useful broader context across multiple model families, end-to-end synthesis pipelines, post-processing, applications, and general challenges, while recent benchmarks~\cite{kindji2025tabular} highlight the importance of tuning budgets and fair evaluation. However, these works are broader in scope or benchmark-oriented rather than dedicated method surveys of diffusion and flow matching models for tabular data. They therefore provide limited analysis of representation choices, feature-type handling,  noising processes, sampling cost, privacy and memorization, and task-specific evaluation across synthesis, imputation, trustworthy generation, and anomaly detection.

To address this gap, we present a comprehensive and critical survey of diffusion and flow matching models for tabular data. To the best of our knowledge, this is the first survey dedicated specifically to this topic. Our main contributions are as follows: 1) We review the historical development of generative models for tabular data and identify the key data-specific challenges that shape diffusion and flow matching methods (Sec.~\ref{sec:ImpHisCha_TabDataGen}); 2) We introduce the necessary preliminaries for diffusion and flow matching models from a tabular perspective, emphasizing the design choices that affect mixed-type data modeling (Sec.~\ref{sec:preliminaries}); 3) We organize and review the literature by application tasks, including synthesis and augmentation, imputation, trustworthy data synthesis, and anomaly detection (Secs.~\ref{sec:DataAug}--\ref{sec:AnoDec}); and 4) We consolidate the lessons from these task-oriented reviews into a discussion of evaluation practice and future directions, including how to compare utility, fidelity, privacy, fairness, sampling cost, and constraint satisfaction; where flow matching has not yet been systematically studied; and what open problems remain in mixed-type representation, feature dependency modeling, relational structure, memorization, and benchmarking (Sec.~\ref{sec:conclusions}). Our coverage spans work from June 2015 to May 2026, with updates maintained in a \href{https://github.com/Diffusion-Model-Leiden/awesome-diffusion-models-for-tabular-data}{GitHub repository}\footnote{\url{https://github.com/Diffusion-Model-Leiden/awesome-diffusion-models-for-tabular-data}}.

\begin{figure*}[t]
	\centering
	\includegraphics[width=1\linewidth]{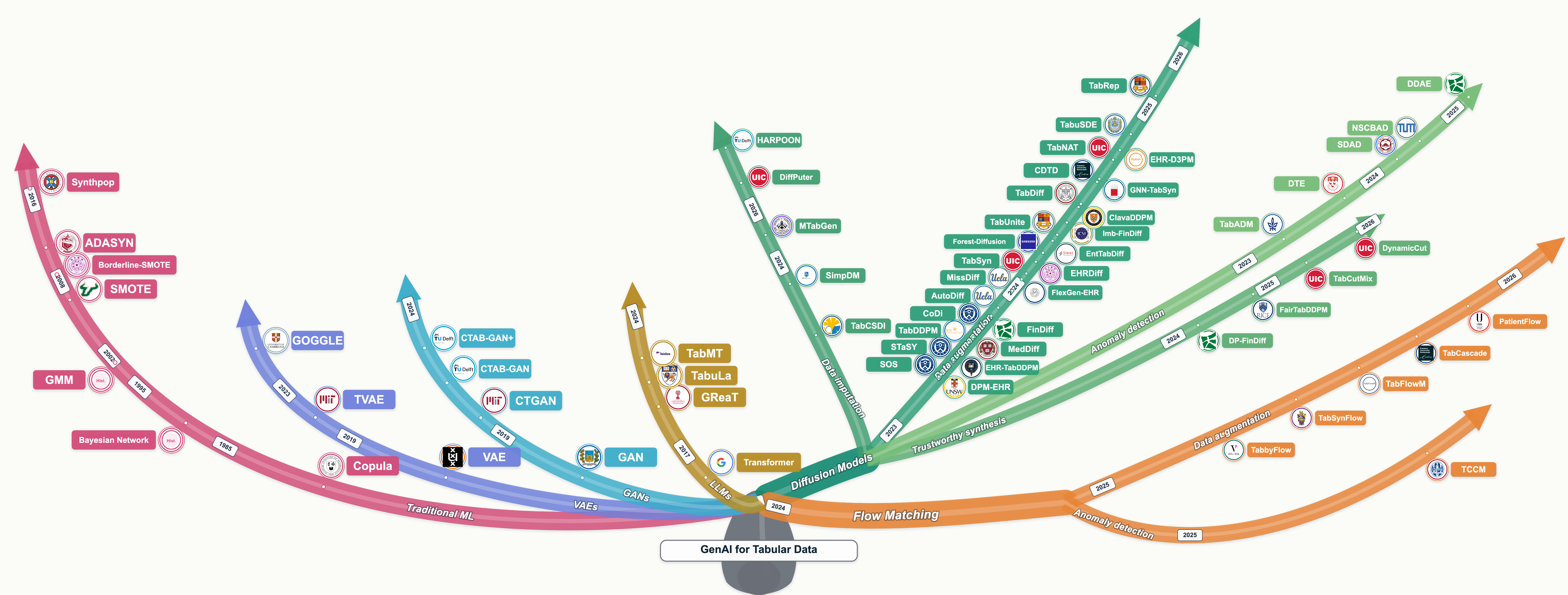}
	\caption{Evolutionary timeline tree of generative models for tabular data. The figure traces six model families from a shared tabular-generation root, with branch-specific chronological markers and method nodes attached to curved branches. Since statistical, VAE-, GAN-, and LLM-based tabular generators serve only as historical context rather than the main focus of this survey, each of these families is represented by a few representative works. In contrast, diffusion and flow matching methods are expanded into task-oriented branches and annotated with first-author affiliation markers when available.}
	\label{fig:TabGenTimeline}
\end{figure*}

\section{Generative Models for Tabular Data}
\label{sec:ImpHisCha_TabDataGen}
This section reviews the history of generative models for tabular data (see Sec.~\ref{subsec:HistoryTabularGen}), then discusses the unique characteristics of tabular data and the challenges these pose for generative models (see Sec.~\ref{Sec:ChallengeswithTabularData}), and finally outlines the application-driven taxonomy of diffusion and flow matching models for tabular data adopted in this survey (see Sec.~\ref{sec:AppOfDMTabular}).

\subsection{History of Generative Models for Tabular Data} 
\label{subsec:HistoryTabularGen}

As shown in Figure~\ref{fig:TabGenTimeline}, prior to the advent of models explicitly designed for data generation, probabilistic models like Copula \cite{sklar1973random}, Gaussian Mixture Models \cite{reynolds2009gaussian} and Bayesian Networks \cite{rabaey2024clinical} were commonly employed for data synthesis. Later, specialized methods, including distance-based approaches such as SMOTE \cite{chawla2002smote} and ADASYN \cite{he2008adasyn}, and probabilistic models like Synthpop \cite{nowok2016synthpop}, were introduced for data synthesis and imputation. However, distance-based methods, including SMOTE, encounter challenges when dealing with large datasets and complex data distributions. Additionally, probabilistic methods, such as Copulas and Synthpop, often struggle with heterogeneous data, impose predefined distributions, and are prone to assumption biases \cite{wang2024challenges}. 

In contrast, deep generative models explicitly designed for data generation have gained increasing prominence in the tabular domain, offering significant advancements and more successful applications compared to traditional tabular data generation techniques. For example, VAE \cite{kingma2019introduction} based methods---such as TVAE \cite{xu2019modeling} and GOGGLE \cite{liu2023goggle}---were shown to achieve superior performance. Importantly, GOGGLE was the first to explicitly model the correlations among features, using a VAE-based model with Graph Neural Networks (GNN) as the encoder and decoder models. However, VAEs-based methods are prone to certain limitations, including blurry outputs in generated data due to the inherent randomness introduced by the latent space and potential difficulty in balancing reconstruction loss and regularization during training, which can affect the quality of the synthetic data. Additionally, VAEs may struggle with accurately capturing multimodal distributions, a common characteristic in real-world tabular datasets.

GAN \cite{goodfellow2014generative} based methods such as CTGAN \cite{xu2019modeling}  have shown promising results for tabular data synthesis. These methods typically use Gaussian Mixture Models to model continuous features, which may be suboptimal for certain real-world data. Additionally, they handle categorical features using one-hot encoding, substantially increasing data dimensionality. Further, GANs suffer from limitations such as mode collapse, failing to capture the full data distribution, and training instability.

LLMs \cite{minaee2024large} have also been explored for tabular data synthesis. For example, GReaT \cite{borisov2023language} converts each row into a natural language sentence and learns sentence-level distributions using GPT \cite{achiam2023gpt}. This approach suffers from some limitations, including potential information loss during data-to-text conversion, increased dimensionality leading to higher computational costs, and context length limitations, which can affect scalability when handling large datasets.

Diffusion models \cite{sohl2015deep} have demonstrated strong performance over VAEs and GANs on image synthesis tasks \cite{dhariwal2021diffusion}. Recent studies, including but not limited to SOS \cite{kim2022sos}, STaSy \cite{kimstasy}, TabDDPM \cite{kotelnikov2023tabddpm}, CoDi \cite{lee2023codi}, and TabSyn \cite{zhangmixed}, suggest that some of these advantages can also transfer to tabular data synthesis. Flow matching has recently become a closely related direction for tabular modeling. Existing studies use it for direct mixed-type synthesis, as in TabbyFlow~\cite{guzman-cordero2025exponential}; latent-space synthesis, as in TabSynFlow~\cite{nasution2026flow} and TabFlowM~\cite{anonymous2026tabflowm}; cascaded coarse-to-fine generation for mixed-type features, as in TabCascade~\cite{mueller2026cascaded}; and longitudinal clinical data generation, as in PatientFlow~\cite{branco2026patientflow}.

%Since this survey focuses specifically on diffusion models for tabular data, we recommend readers refer to dedicated survey papers \cite{wang2024challenges, kim2024generative,fang2024large} for a comprehensive review of other generative models for tabular data.

%[@Zhong: I will add more descriptions for each representative method after reading some related work.] 
 
\subsection{Challenges with Generative Models for Tabular Data}
\label{Sec:ChallengeswithTabularData}
Training generative models for tabular data is especially difficult due to the following seven data-specific challenges.

\subsubsection{Missing Values} This phenomenon often happens in real-world tabular datasets for several reasons \cite{zheng2022diffusion}, e.g., privacy concerns (people's refusal to answer
questions about their employment or income information in census data \cite{lillard1986we}), difficulty of data collection (drop-out in studies and merging unrelated data in healthcare data \cite{wells2013strategies}), human operation errors when processing data \cite{emmanuel2021survey}, or machine error due to malfunctioning of equipment \cite{emmanuel2021survey}. Values can be Missing At Random (MAR), Missing Completely At Random (MCAR), or Missing Not At Random (MNAR) \cite{rubin1976inference}. Most generative models for tabular data cannot be directly trained on incomplete data \cite{jolicoeur2024generating}. 

\subsubsection{Intricate Individual Feature Distribution} A feature in tabular data, i.e., a column, may have a `complicated distribution' \cite{xu2019modeling}: 2.1) for a numerical feature, the distribution can be non-Gaussian and/or it can have multiple distribution modes; 2.2) for a categorical feature, the class distribution can be highly imbalanced; or 2.3) within the same feature type (numerical or categorical), different features usually have different statistical properties (e.g., feature-wise marginal distribution) due to the fact that the meanings of features can be different. %In contrast, pixel values of each image in an image dataset are usually assumed to follow the same distribution.

\subsubsection{Heterogeneous Features} Heterogeneous features refer to columns that contain different types of data, which has been considered the most challenging issue \cite{suh2023autodiff}. These types may include numerical, categorical, ordinal, boolean, text, or datetime data. This diversity makes modeling such data more complex than datasets consisting of only one feature type. It is unclear whether it is effective to combine different types of models, that were designed for different types of features.

\subsubsection{Feature Dependencies (Correlations between Features)} Tabular data synthesis models should learn the joint probability of multiple columns, but these columns are usually not independent of each other (i.e., their joint probability cannot be decomposed into the product of their marginal probabilities). Unlike image data, which contains only continuous pixel values with local spatial correlations, or text data, which comprises tokens sharing the same dictionary space \cite{shi2024tabdiff}, capturing correlations among features has been a long-standing challenge~\cite{suh2023autodiff}. The heterogeneous nature of tabular data makes it even more challenging than capturing the correlations among purely categorical or numerical features.

\subsubsection{Mixed-Type Feature} A mixed-type feature refers to a single feature  that contains more than one type of data. Unlike heterogeneous features, which describe the diversity of feature types across columns, mixed-type features involve a single column containing values of different data types, such as numerical and categorical data mixed together.

\subsubsection{Small Data Size} Compared to image or text datasets, which typically have massive amounts of available data, tabular datasets are usually smaller in volume. Specifically, the number of samples (or rows) is often limited, making it challenging to effectively train deep models.

\subsubsection{Domain-specific Constraints} Real-world tabular datasets span various domains, such as healthcare, finance, and engineering, each with domain-specific features and constraints. For example, healthcare data must maintain realistic ranges for age or blood pressure, while finance data requires valid transaction amounts and balances. These constraints are often difficult to enforce or even verify.

\subsection{Taxonomy of Diffusion and Flow Matching Models for Tabular Data }
 \label{sec:AppOfDMTabular}

 \begin{figure}[t]
    \centering
    \includegraphics[width=\columnwidth]{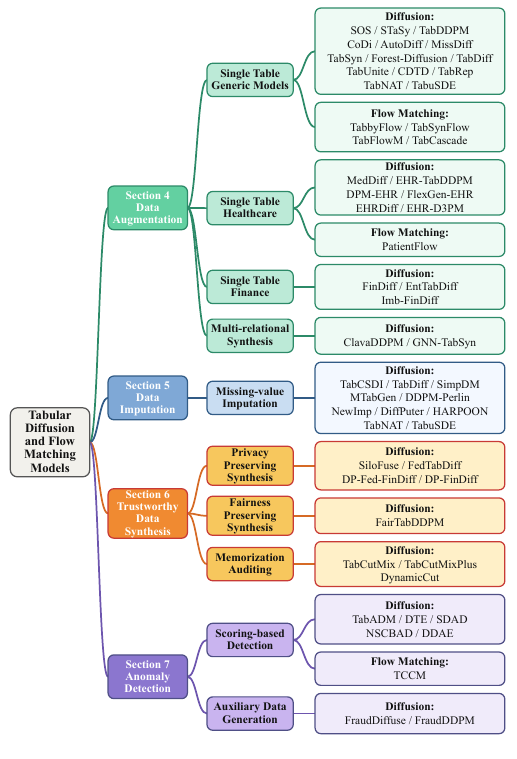}
    \caption{Taxonomy of diffusion and flow matching models for tabular data. The taxonomy is organized by survey section and task category, grouping representative methods by application role rather than by publication time.}
    \label{fig:taxonomy}
\end{figure}

On a high level, generative models for tabular data are generally designed for one of four key tasks: data augmentation (see Sec.~\ref{sec:DataAug}), data imputation (see Sec.~\ref{sec:DataImp}), trustworthy data synthesis (see Sec.~\ref{sec:TrustSynth}), and anomaly detection (see Sec.~\ref{sec:AnoDec}). While this applies to tabular generative models in general, the focus of this survey is on diffusion and flow matching models. Therefore, we systematically examine diffusion- and, where available, flow-matching-based approaches for each task. This application-driven taxonomy was chosen instead of a classification based only on the underlying generative model types (see Sec.~\ref{sec:preliminaries}) because it provides a more intuitive and practical perspective on how these models are applied to real-world problems. The taxonomy, summarized in Figure~\ref{fig:taxonomy}, allows us to systematically categorize and evaluate the literature on diffusion and flow matching models for tabular data. 

Sec.~\ref{sec:DataAug} reviews \textit{data augmentation} methods that generate synthetic data, including single-table synthesis (see Sec.~\ref{subsec:DataAug_Single}) and multi-relational synthesis (see Sec.~\ref{subsec:DataAug_Multi}). Then, Sec.~\ref{sec:DataImp} discusses \textit{data imputation} approaches focused on handling missing values in tabular datasets. Next, Sec.~\ref{sec:TrustSynth} explores \textit{trustworthy data synthesis}, which includes privacy-preserving, fairness-preserving, and memorization-aware data generation techniques. Finally, Sec.~\ref{sec:AnoDec} reviews \textit{anomaly detection} methods that leverage diffusion or flow matching models to identify anomalies by learning the normal data distribution or a normal-data transport field~\cite{li2025scalable}. The absence of dedicated flow-matching methods in some branches is itself informative: it identifies task settings where flow-matching methods for tabular data have not yet been systematically studied, rather than implying that diffusion and flow matching are equally represented in every section.

%Each of Secs.~\ref{sec:DataAug}--\ref{sec:AnoDec} begins with a brief introduction, accompanied by a summary table of the relevant works. This is followed by an in-depth chronological review of the papers, emphasizing how they address the key challenges, along with their performance evaluations and limitations. 
For completeness, we next introduce diffusion and flow matching models and the prominent model types used for tabular data in Sec.~\ref{sec:preliminaries}. Readers familiar with these concepts may skip this section.

\section{Diffusion and Flow Matching Models Preliminaries}
\label{sec:preliminaries}

This section first introduces the core mechanism of diffusion models, covering both the forward diffusion process (i.e., gradual noise addition) and the reverse process (i.e., learning to denoise). We then present prominent diffusion model types, including Gaussian diffusion models (DDPMs) (see Sec.~\ref{sec:Background_Math_DDPM}), multinomial diffusion models (see Sec.~\ref{sec:Background_Math_Multinomial}), score-based generative models (SGMs) (see Sec.~\ref{sec:Background_Math_Score}), score-based generative models through stochastic differential equations (SDEs) (see Sec.~\ref{sec:Background_Math_ScoreSDEs}), and conditional diffusion models (see Sec.~\ref{sec:Background_Math_ConDiffusion}). Finally, we introduce flow matching (see Sec.~\ref{sec:Background_Math_FlowMatching}), which replaces explicit denoising or score estimation with learning a time-dependent velocity field that transports samples from a base distribution to the data distribution. Throughout this survey, bold lowercase letters denote vectors, bold uppercase letters denote matrices, and task-specific notation is introduced locally where each problem is defined.

% \subsection{Mathematics of Diffusion Models}
% \label{sec:Background_Math_DM} 

\textit{Diffusion probabilistic models}, more commonly known as \textit{diffusion models} \cite{sohl2015deep}, are deep generative models defined from a forward diffusion process and a reverse denoising process. First, the \textit{diffusion process} aims to gradually `corrupt' a sample $\mathbf{x}_{0}$ (drawn from training data distribution $q_{\text{data}}(\cdot)$) to a noisy instance $\mathbf{x}_{T}$ (from a prior distribution $q_{\text{noise}}(\cdot)$) using
\begin{equation}
    q(\mathbf{x}_{1:T}\vert\mathbf{x}_{0}) := \prod_{t=1}^{T}q(\mathbf{x}_{t}\vert \mathbf{x}_{t-1}),\label{equ:DDPMForward}
\end{equation}
with $q(\mathbf{x}_{t}\vert \mathbf{x}_{t-1})$ the forward transition probability. Second, the \textit{denoising process} attempts to remove noise and generate a synthetic but realistic sample $\hat{\mathbf{x}}_{0}$ from $\mathbf{x}_{T}$ using
\begin{equation}
    p_{\boldsymbol{\theta}}(\mathbf{x}_{0:T}):= p(\mathbf{x}_{T})\prod_{t=1}^{T}p_{\boldsymbol{\theta}}(\mathbf{x}_{t-1}\vert \mathbf{x}_{t}),\label{equ:DDPMReverse}
\end{equation}
with $p_{\boldsymbol{\theta}}(\mathbf{x}_{t-1}\vert \mathbf{x}_{t})$ an approximation of the reverse of the forward transition probability, which is often learned by a neural network with parameters $\boldsymbol{\theta}$. Particularly, they learn $\boldsymbol{\theta}$ by minimizing the following variational upper bound ($L_{\text{VUB}}$) on the negative log-likelihood:

\begin{align}
   - \log \,\, & p(\mathbf{x}) \leq \mathbb{E}_{q(\mathbf{x}_{1}|\mathbf{x}_{0})}[   \underbrace{-\log p_{\boldsymbol{\theta}}(\mathbf{x}_{0}\vert \mathbf{x}_{1})}_{L_0(L_{\text{reconstruction}})} \nonumber ] \\ 
   & + \underbrace{D_{KL}[q(\mathbf{x}_{T}\vert\mathbf{x}_{0})\vert\vert p(\mathbf{x}_{T})]}_{L_T(L_{\text{prior}})} \nonumber \\
    & + \sum_{t=2}^{T} \mathbb{E}_{q(\mathbf{x}_{t}|\mathbf{x}_{0})} \underbrace{D_{KL}[q(\mathbf{x}_{t-1}\vert\mathbf{x}_{t},\mathbf{x}_{0})\vert \vert p_{\boldsymbol{\theta}}(\mathbf{x}_{t-1}\vert \mathbf{x}_{t})]}_{L_{t}(L_{\text{diffusion}})} .
    \label{equ:DDPMLoss}
\end{align}

Here, $L_0$ can be interpreted as a reconstruction term that predicts the log probability of original sample $\mathbf{x}_{0}$ given the noised latent $\mathbf{x}_{1}$, while $L_T$ measures how the final corrupted sample $\mathbf{x}_{T}$ resembles the noise prior distribution. Meanwhile, $L_{t}$ measures how close the estimated $p_{\boldsymbol{\theta}}(\mathbf{x}_{t-1}\vert \mathbf{x}_{t})$ is to the ground-truth posterior transition probability $q(\mathbf{x}_{t-1}\vert\mathbf{x}_{t},\mathbf{x}_{0})$.

The formulations above only define the generic structure of diffusion models: depending on the specific data type (namely continuous or discrete), the definitions of prior noise distribution $p(\mathbf{x}_{T})$, forward transition probability $q(\mathbf{x}_{t} \vert \mathbf{x}_{t-1})$, the reverse of forward transition probability $p_{\theta}(\mathbf{x}_{t-1}\vert \mathbf{x}_{t})$, and their training objectives can be different. In the following, we concisely review the five diffusion model frameworks most commonly used for tabular data modeling.

\subsubsection{Gaussian Diffusion Models} \label{sec:Background_Math_DDPM} 
Gaussian diffusion models, often known as Denoising Diffusion Probabilistic Models (DDPMs) \cite{ho2020denoising}, operate in continuous spaces. They
define the diffusion and reverse processes as
\begin{subequations} \label{equ:DDPMEquationsCon}
\begin{align}
    p(\mathbf{x}_{T}) & := \mathcal{N}(\mathbf{x}_{T};\mathbf{0},\mathbf{I}), 
    \label{equ:DDPMContinuousPrior} \\
    q(\mathbf{x}_{t}\vert \mathbf{x}_{t-1}) & := \mathcal{N}(\mathbf{x}_{t};\sqrt{1-\beta_{t}}\mathbf{x}_{t-1},\beta_{t}\mathbf{I}),
    \label{equ:DDPMContinuousForward} \\
    p_{\boldsymbol{\theta}}(\mathbf{x}_{t-1}\vert \mathbf{x}_{t}) & := \mathcal{N}(\mathbf{x}_{t-1};\boldsymbol{\mu}_{\boldsymbol{\theta}}(\mathbf{x}_{t},t),\boldsymbol{\Sigma}_{\boldsymbol{\theta}}(\mathbf{x}_{t},t)),
    \label{equ:DDPMContinuousBackward}
\end{align}
\end{subequations}
where Gaussian noises are gradually injected into the sample based on a time-dependent variance schedule $\{\beta_{t}\}_{t=1}^{T}$, with $\beta_{t}\in(0,1)$ determining the amount of noise added at time step $t$. To approximate $p_{\boldsymbol{\theta}}(\mathbf{x}_{t-1}\vert \mathbf{x}_{t})$, \cite{ho2020denoising} define
\begin{equation}
\begin{aligned}
\label{eq:DDPM_mu_Sigma}
    \boldsymbol{\mu}_{\boldsymbol{\theta}}(\mathbf{x}_{t},t) &= \frac{1}{\sqrt{\alpha_{t}}}\Big(\mathbf{x}_{t} - \frac{\beta_{t}}{\sqrt{1-\bar{\alpha}_{t}}}\boldsymbol{\epsilon}_{\boldsymbol{\theta}}(\mathbf{x}_{t},t)\Big), \\
    \boldsymbol{\Sigma}_{\boldsymbol{\theta}}(\mathbf{x}_{t},t) &= \sigma_{t}\mathbf{I}, 
\end{aligned}
\end{equation}
where $\sigma_{t}$ controls the noise level added at time step $t$, $\alpha_{t} := 1- \beta_{t},\bar{\alpha}_{t}:=\prod_{i=1}^{t}\alpha_{i}$, and $\boldsymbol{\epsilon}_{\boldsymbol{\theta}}$ is a neural network to predict ground truth noise $\boldsymbol{\epsilon} \sim \mathcal{N}(\mathbf{0},\mathbf{I})$ that has been added to noise sample $\mathbf{x}_{t}$. As a result, they propose to optimize the following simplified objective function (rather than Eq.~\ref{equ:DDPMLoss}):
\begin{equation}
    L_{\text{simple}}^{\text{Gauss}}(\boldsymbol{\theta}) := \mathbb{E}_{t}\mathbb{E}_{\mathbf{x}_{0} \sim q(\mathbf{x}_{0})}\mathbb{E}_{\boldsymbol{\epsilon} \sim \mathcal{N}(\mathbf{0},\mathbf{I})}\big[\lambda(t)\Vert\boldsymbol{\epsilon}-\boldsymbol{\epsilon}_{\boldsymbol{\theta}}(\mathbf{x}_{t},t)\Vert_{2}^{2}\big],\label{equ:DDPMContinuousLossSimple}
\end{equation}
where $\lambda(t)$ is a weighting function to adjust the noise scales, and $\Vert \cdot\Vert_2$ denotes the Euclidean norm. %Further note that $\mathbf{x}_t$ is a function of $\mathbf{x}_0$ and hence we do not need to explicitly take the expectation with respect to $\mathbf{x}_t$.

\subsubsection{Multinomial Diffusion Models} \label{sec:Background_Math_Multinomial} 
Multinomial diffusion models \cite{hoogeboom2021argmax} \cite{austin2021structured} operate in discrete spaces, designed to generate categorical data $\mathbf{x}_{t} \in \{0,1\}^{K}$ (i.e., one-hot encoding with $K$ distinct values). They are defined as follows:
\begin{subequations} \label{equ:DDPMEquationsDis}
\begin{align}
    p(\mathbf{x}_{T}) & := \mathcal{\text{Cat}}(\mathbf{x}_{T};1/K), 
    \label{equ:DDPMDiscretePrior} \\
    q(\mathbf{x}_{t}\vert \mathbf{x}_{t-1}) & := \mathcal{\text{Cat}}(\mathbf{x}_{t};(1-\beta_{t})\mathbf{x}_{t-1}+\beta_{t}/K),
    \label{equ:DDPMDiscreteForward} \\
    p_{\boldsymbol{\theta}}(\mathbf{x}_{t-1}\vert \mathbf{x}_{t}) & := \sum_{\hat{\mathbf{x}}_{0}=1}^{K}q(\mathbf{x}_{t-1}\vert \mathbf{x}_{t},\hat{\mathbf{x}}_{0})p_{\boldsymbol{\theta}}(\hat{\mathbf{x}}_{0}\vert\mathbf{x}_{t}),
    \label{equ:DDPMDiscreteBackward}
\end{align}
\end{subequations}
with $\mathcal{\text{Cat}}(\cdot)$ categorical distribution and $K$ the number of categories (the computation between scalars and vectors are done in an element-wise way). Importantly, \textit{uniform noise} (rather than Gaussian noise) is added to the sample according to the noise schedule $\beta_{t}$. As we can see,  $p_{\boldsymbol{\theta}}(\mathbf{x}_{t-1}\vert \mathbf{x}_{t})$ is parameterized as $q(\mathbf{x}_{t-1}\vert \mathbf{x}_{t},\hat{\mathbf{x}}_{0}(\mathbf{x}_{t},t))$, with $\hat{\mathbf{x}}_{0}(\mathbf{x}_{t},t)$ predicted by a neural network, which can be trained via the multinomial diffusion loss  defined using Eq.~\ref{equ:DDPMLoss}.
% As a result, the multinomial diffusion loss $L^{\text{Mult}}$ is still defined using Eq.~\ref{equ:DDPMLoss}.
Note that this model can handle only one categorical feature at a time. In other words, for a table with \( C \) categorical features, \( C \) separate multinomial diffusion models would need to be built. %(DDIM \cite{song2021denoising} may offer a more efficient solution.) 
%More importantly, several novel diffusion models for discrete data have been proposed recently; these will be reviewed in Sec.~\ref{subsec:DiffModelInDiscreteAll} for better readability.

\subsubsection{Score-based Generative Models (SGMs)}
\label{sec:Background_Math_Score} For these models, the forward diffusion process follows the same structure as the Gaussian diffusion model, whereas the reverse process is defined differently, as shown below. Given an instance $\mathbf{x}$ and its distribution $p(\mathbf{x})$, its score function is defined as $\nabla_{\mathbf{x}}\log p(\mathbf{x})$. To estimate the score function, one can train a neural network $S_{\boldsymbol{\theta}}(\cdot)$ with the following objective:
\begin{equation}
    \mathbb{E}_{\mathbf{x}\sim p(\mathbf{x})}\Vert S_{\boldsymbol{\theta}}(\mathbf{x}) - \nabla_{\mathbf{x}}\log p(\mathbf{x}) \Vert^{2}_{2}.
\end{equation}
However, Song \& Ermon \cite{song2019generative} point out that the estimated score functions are inevitably imprecise in low density regions when the low-dimensional manifolds are embedded into a high-dimensional space. To mitigate this, in the diffusion process they perturb the original data $\mathbf{x}$ with a sequence of random
Gaussian noises with intensifying scales $0<\sigma_{1}<\cdots <\sigma_{T}$. In other words, $p_{\sigma_{1}}\approx p(\mathbf{x}_{0})$, $p_{\sigma_{T}}\approx \mathcal{N}(\mathbf{0},\mathbf{I})$, and $p_{\sigma_{t}}\approx \mathcal{N}(\mathbf{x}_{t};\mathbf{x}_{0},\sigma_{t}^{2}\mathbf{I})$. In the reverse process, they utilize a noise-conditioned score network $S_{\boldsymbol{\theta}}(\cdot)$ to approximate $\nabla_{\mathbf{x}}\log p_{\sigma_{t}}(\mathbf{x})$, which analytically equals $(\mathbf{x}_{0}-\mathbf{x}_{t})/\sigma_{t}$. As a result, the training objective is as follows:
\begin{equation}
    \frac{1}T{}\sum_{t=1}^{T}\lambda(\sigma_{t})\mathbb{E}_{p(\mathbf{x}_{0})}\mathbb{E}_{\mathbf{x}_{t}\sim p_{\sigma_{t}}(\mathbf{x}_{t}\vert\mathbf{x})}\left\Vert S_{\boldsymbol{\theta}}(\mathbf{x}_{t},\sigma_{t})+(\frac{\mathbf{x}_{t}-\mathbf{x}_{0}}{\sigma_{t}})\right\Vert^{2}_{2}.
\end{equation}
After training $ S_{\boldsymbol{\theta}}(\cdot)$, new samples are generated with the annealed Langevin dynamics (see \cite{song2019generative} for details).% Note that SGMs are defined in discrete time space, a special case corresponding to the variance exploding form in the generalized version presented in the sequel.

\subsubsection{Score-based Generative Models through Stochastic Differential Equations (SDEs) \cite{songscore}} \label{sec:Background_Math_ScoreSDEs} This is a continuous-time generalization of denoising diffusion models and score based generative models. Particularly, the \textit{diffusion process} is defined with the following stochastic differential equation \cite{kloeden1992stochastic}:
\begin{equation}
    d\mathbf{x} = \mathbf{f}(\mathbf{x},t)dt +g(t)d\mathbf{w}, \label{equ:ScoreSDEForward}
\end{equation}
where $\mathbf{f}(\mathbf{x},t)=f(t)\mathbf{x}$, and $f(\cdot)$, $g(\cdot)$ are referred to as the drift and diffusion coefficients of $\mathbf{x}_{t}$, respectively. 
%where $\mathbf{f}(\mathbf{x},t)=f(t)\mathbf{x}$, and $f(\cdot)$, $g(\cdot)$ are referred to as the drift and diffusion coefficients of $\mathbf{x}_{t}$, respectively. Concretely, $\mathbf{x}_{t}$ is the value of sample $\mathbf{x}$ at time point $t$, where $\mathbf{x}_{0}$ is a real sample and $\mathbf{x}_{T}$ is a noised sample. 
Moreover, $\mathbf{w}$ is the standard Wiener process. The most widely studied and commonly used diffusion models can be broadly categorized into three main types: 1) Variance Exploding (VE), 2) Variance Preserving (VP), and 3) sub-Variance Preserving (sub-VP) based on the types of functions  $f(\cdot)$ and $g(\cdot)$ as follows:
\begin{align}
    \mathbf{f}(\mathbf{x},t)&=\begin{cases}0,  &\textrm{ if VE,}\\
    -\frac{1}{2}\gamma_{t}\mathbf{x},&\textrm{ if VP,}\\
    -\frac{1}{2}\gamma_{t}\mathbf{x},&\textrm{ if sub-VP,}\\
    \end{cases}\label{equ:ScoreSDEdrift}\\
    g(t)&=\begin{cases}\sqrt{\frac{\texttt{d}[\sigma^{2}_{t}]}{\texttt{d}t}},&\textrm{ if VE,}\\
    \sqrt{\gamma_{t}},&\textrm{ if VP,}\\
    \sqrt{\gamma_{t}(1-e^{-2\int_0^t \gamma_{s}\, \texttt{d}s})},&\textrm{ if sub-VP,}\\
    \end{cases}\label{equ:dScoreSDEiffusion}
\end{align}
where $\gamma_{t}$ and $\sigma_{t}$ are noise functions w.r.t. the time variable $t$. Meanwhile, the \textit{denoising process} is defined as the reverse of the diffusion process:
\begin{equation}
    d\mathbf{x}=[\mathbf{f}(\mathbf{x},t)-g(t)^{2}\nabla_{\mathbf{x}}\log p_{t}(\mathbf{\mathbf{x}})]dt+g(t)d\bar{\mathbf{w}},\label{equ:ScoreSDEReverse}
\end{equation}
where $\bar{\mathbf{w}}$ is a Wiener process running backward in time, and the score function $\nabla_{\mathbf{x}}\log p_{t}(\mathbf{\mathbf{x}})$ is approximated by a learnable neural network $S_{\boldsymbol{\theta}}(\mathbf{x},t)$.  However, directly approximating the score function is computationally intractable and thus they propose to  train $S_{\boldsymbol{\theta}}(\cdot)$ by estimating the transition probability $\nabla_{\mathbf{x}_{t}}\log p(\mathbf{\mathbf{x}}_{t}\vert\mathbf{x}_{0})$ as follows \cite{songscore}:
\begin{align}
    \underset{\theta}{\arg\min} & \, \mathbb{E}_{t}\Big\{\lambda(t) \mathbb{E}_{\mathbf{x}_{0}} \Big[\mathbb{E}_{\mathbf{x}_{t}\vert\mathbf{x}_{0}} \Big[ \nonumber \\
    & \lVert S_{\boldsymbol{\theta}}(\mathbf{x}_{t},t) - \nabla_{\mathbf{x}_t} \log p(\mathbf{x}_t \vert \mathbf{x}_0) \rVert^{2}_{2} \Big]\Big]\Big\},
    \label{equ:ScoreSDELoss}
\end{align}
where $\nabla_{\mathbf{x}_{t}}\log p(\mathbf{\mathbf{x}}_{t}\vert\mathbf{x}_{0})$ follows the Gaussian distribution and can be collected during the diffusion process. Moreover, $\lambda(t)$ is used to trade-off between sample quality and likelihood. 

After training $S_{\boldsymbol{\theta}}(\cdot)$, we can generate new samples with either the 
\textit{predictor-corrector} or the \textit{probability flow} framework. In general, the latter is preferred due to its fast sampling and exact log-probability computation compatibilities (see \cite{songscore}). 
%In short, the \textit{probability flow} employs the following neural ordinary differential equation (NODE) based model \cite{chen2018neural}:
%\begin{equation}
%    d\mathbf{x} = \left(\mathbf{f}(\mathbf{x},t)-\frac{1}{2}g(t)^{2}\nabla_{\mathbf{x}}\log p_{t}(\mathbf{x}) \right)dt, \label{equ:ScoreSDE_NODESolver}
%\end{equation}
%which describes a deterministic process whose marginal probability is equivalent to that of the original reverse SDE (namely Eq.~\ref{equ:ScoreSDEReverse}) \cite{songscore}.

\subsubsection{Conditional Diffusion Models}\label{sec:Background_Math_ConDiffusion}
The diffusion models introduced so far are all unconditional, meaning the posterior estimator function \( p_{\boldsymbol{\theta}}(\cdot) \) does not know the label of the data it is modeling. Given a label vector $\mathbf{y}$, \cite{sohl2015deep} and \cite{dhariwal2021diffusion} suggest that this can be achieved through a so-called \textit{conditional reverse process} in DDPM, defined as:
\begin{equation}
\label{Equ:ClassifierGuidanceDDPM}
    p_{\boldsymbol{\theta},\boldsymbol{\phi}}(\mathbf{x}_{t-1}\vert \mathbf{x}_{t},\mathbf{y}) \propto p_{\boldsymbol{\theta}}(\mathbf{x}_{t-1}\vert \mathbf{x}_{t})p_{\boldsymbol{\phi}}(\mathbf{y}\vert \mathbf{x}_{t-1}),
\end{equation}
which requires to train a classifier $p_{\boldsymbol{\phi}}(\cdot)$ and thus is known as \textit{classifier-guided} DDPM. Dhariwa and Nichol \cite{dhariwal2021diffusion} further approximate the logarithm of it with a perturbed Gaussian transition as follows:
\begin{equation}
\label{Equ:ClassifierGuidanceApprDDPM}
    \log( p_{\boldsymbol{\theta},\boldsymbol{\phi}}(\mathbf{x}_{t-1}\vert \mathbf{x}_{t},\mathbf{y})) \approx \log(p(\mathbf{z}))+C,
\end{equation}
where $C$ is a constant, $\mathbf{z} \sim \mathcal{N}(\boldsymbol{\mu} + \boldsymbol{\Sigma}\mathbf{g},\boldsymbol{\Sigma})$, and $\mathbf{g} = \nabla_{\mathbf{x}_{t-1}} \log (p_{\boldsymbol{\phi}}(\mathbf{y}\vert \mathbf{x}_{t-1}))\lvert_{\mathbf{x}_{t-1} =\boldsymbol{\mu}}$ is computed from the classifier $p_{\boldsymbol{\phi}}(\cdot)$. To avoid training a separate classifier, Ho \& Salimans \cite{ho2021classifier} propose a \textit{classifier-free guided} DDPM:
\begin{equation}
\label{Equ:ClassifierFreeGuidanceDDPM}
    \bar{\boldsymbol{\epsilon}}(\mathbf{x}_{t},\mathbf{y},t) = \hat{\boldsymbol{\epsilon}}(\mathbf{x}_{t},t)+\omega_{g}[\hat{\boldsymbol{\epsilon}}(\mathbf{x}_{t},\mathbf{y},t)-\hat{\boldsymbol{\epsilon}}(\mathbf{x}_{t},t)],
\end{equation}
where $\hat{\boldsymbol{\epsilon}}(\cdot)$ is the DDPM noise estimator $\boldsymbol{\epsilon}_{\theta}(\cdot)$,  $[\hat{\boldsymbol{\epsilon}}(\mathbf{x}_{t},\mathbf{y},t)-\hat{\boldsymbol{\epsilon}}(\mathbf{x}_{t},t)]$ is guidance of $\mathbf{y}$, and $\omega_{g}$ the guidance weight.

\subsubsection{Flow Matching Models}\label{sec:Background_Math_FlowMatching}
Flow matching~\cite{lipman2023flow} is a continuous-time generative modeling framework that learns a velocity field rather than a denoising network or score function. Let $p_{0}$ be a simple base distribution, such as a standard Gaussian, and let $p_{1}=p_{\text{data}}$ be the target data distribution. A time-dependent flow map $\phi_{t}$ transports samples from $p_{0}$ to $p_{1}$ according to the ordinary differential equation
\begin{equation}
    \frac{d}{dt}\phi_{t}(\mathbf{x})=\mathbf{v}_{\boldsymbol{\theta}}(\phi_{t}(\mathbf{x}),t), \quad \phi_{0}(\mathbf{x})=\mathbf{x}.
    \label{equ:FlowMatchingODE}
\end{equation}
Training can be formulated as velocity regression along a chosen probability path. For example, with a linear interpolation between a noise sample $\mathbf{x}_{0}\sim p_{0}$ and a data sample $\mathbf{x}_{1}\sim p_{1}$, one can define
\begin{equation}
    \mathbf{x}_{t}=(1-t)\mathbf{x}_{0}+t\mathbf{x}_{1}, \quad \mathbf{u}_{t}(\mathbf{x}_{t})=\mathbf{x}_{1}-\mathbf{x}_{0},
    \label{equ:FlowMatchingPath}
\end{equation}
and optimize
\begin{equation}
    \mathcal{L}_{\text{FM}}(\boldsymbol{\theta})=\mathbb{E}_{t,\mathbf{x}_{0},\mathbf{x}_{1}}\left[\left\Vert \mathbf{v}_{\boldsymbol{\theta}}(\mathbf{x}_{t},t)-\mathbf{u}_{t}(\mathbf{x}_{t})\right\Vert_{2}^{2}\right].
    \label{equ:FlowMatchingLoss}
\end{equation}
Unlike diffusion models, which typically specify a forward noising process and learn to reverse it through denoising or score estimation, flow matching directly learns a velocity field defined along a probability path. This distinction is important for tabular data because the choice of probability path, representation space, sampling procedure, and feature-type parameterization can affect mixed-type feature handling, sampling cost, privacy risk, and the preservation of cross-feature dependencies. We therefore use these axes to compare diffusion and flow matching methods in later sections, rather than treating flow matching only as a faster sampler.

\begin{figure}[tbp]
\centering
\includegraphics[width=\columnwidth]{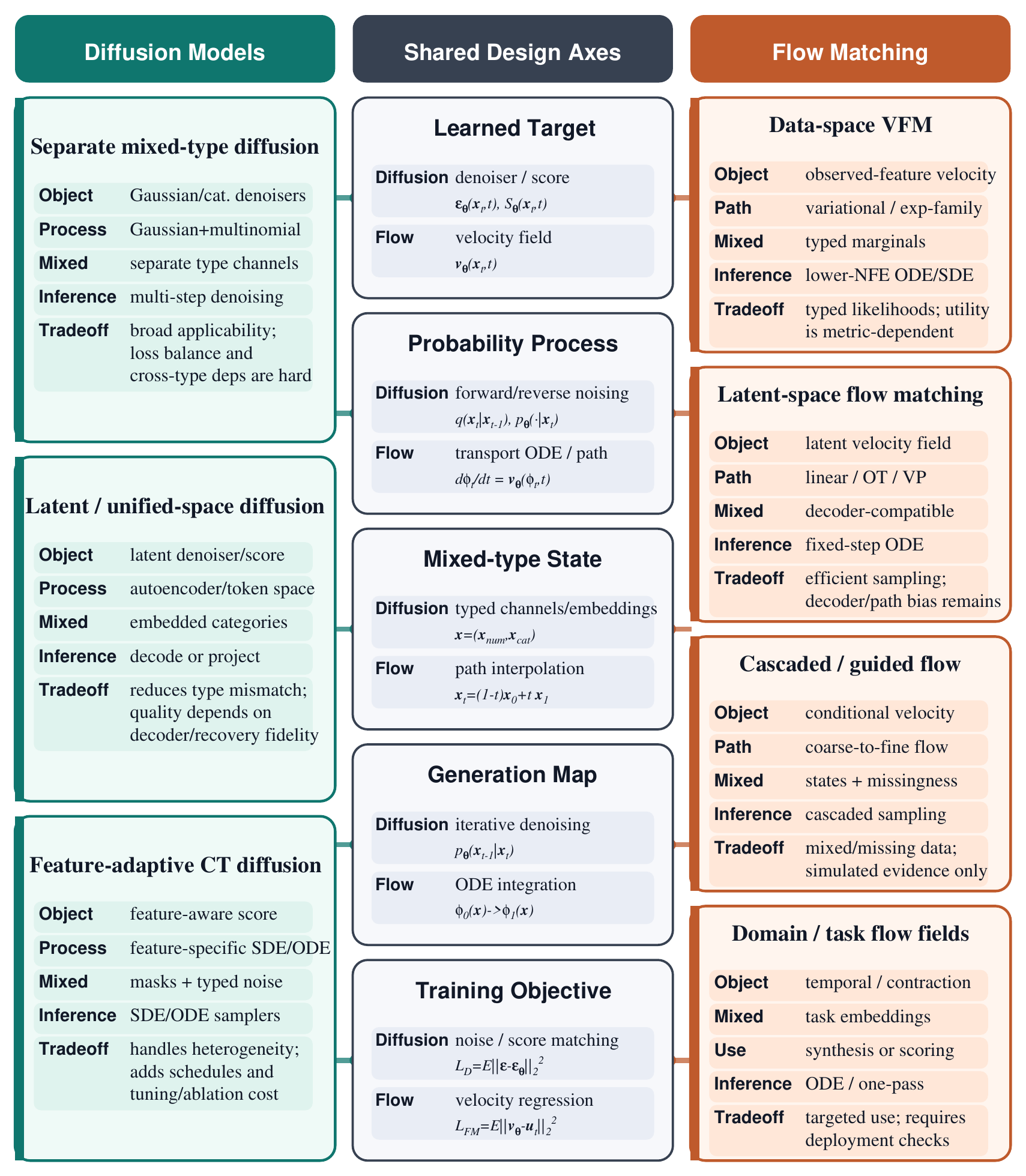}
\caption{Design-axis map for tabular diffusion and flow matching models. The left and right columns summarize representative diffusion and flow-matching routes, respectively, while the middle column aligns them through shared mathematical axes: learned target, probability process, mixed-type state, generation map, and training objective. The connectors emphasize that diffusion and flow matching make different modeling choices along comparable tabular design dimensions.}
\label{fig:DesignAxesDiffusionFlow}
\end{figure}

Figure~\ref{fig:DesignAxesDiffusionFlow} translates the mathematical preliminaries above into a compact reading map for the remainder of this survey. Diffusion models differ in whether they use separate mixed-type corruption processes, latent or unified continuous representations, or feature-adaptive continuous-time dynamics. Flow matching models, in turn, differ in whether the velocity field is learned in observed data space, a latent/token space, a cascaded coarse-to-fine space, or a task-specific space. The shared middle column makes these routes directly comparable: diffusion typically learns a denoiser, score, or reverse kernel such as $\boldsymbol{\epsilon}_{\boldsymbol{\theta}}(\mathbf{x}_{t},t)$, $S_{\boldsymbol{\theta}}(\mathbf{x}_{t},t)$, or $p_{\boldsymbol{\theta}}(\mathbf{x}_{t-1}\vert\mathbf{x}_{t})$, whereas flow matching learns a velocity field $\mathbf{v}_{\boldsymbol{\theta}}(\mathbf{x}_{t},t)$ and generates by integrating a flow map $\phi_{t}$ as in Eq.~\ref{equ:FlowMatchingODE}. This comparison is not merely about sampling speed; it also affects how methods represent mixed numerical--categorical states, choose probability paths or noising schedules, balance feature-type losses, and expose utility--privacy--cost tradeoffs. Accordingly, the following sections evaluate methods by both application task and these shared design axes across data augmentation (see Sec.~\ref{sec:DataAug}), data imputation (Sec.~\ref{sec:DataImp}), trustworthy data synthesis (Sec.~\ref{sec:TrustSynth}), and anomaly detection (Sec.~\ref{sec:AnoDec}), rather than treating all synthetic data generators as interchangeable.
 
\section{Diffusion and Flow Matching Models for Data Augmentation}
\label{sec:DataAug}

\textit{Data augmentation} for tabular data is a long-standing research problem  \cite{cui2024tabular}. In general, it can be divided into two different tasks: 1) \textit{data synthesis}, which is the process of generating synthetic data that mimics the characteristics of real-world data, and can be further divided into \textit{single table synthesis} and \textit{multi-relational dataset synthesis}; and 2) \textit{over-sampling}, which balances an imbalanced table by increasing the number of samples in the minority class(es). Particularly, {over-sampling} can be considered as a special case of \textit{single table synthesis} where we only generate a part of the table. Accordingly, in this survey, we categorize relevant works into \textit{diffusion and flow matching models for single-table synthesis} (including over-sampling)  and \textit{diffusion models for multi-relational dataset synthesis} , providing their formal definitions and reviewing related studies in Secs.~\ref{subsec:DataAug_Single} and~\ref{subsec:DataAug_Multi}, respectively.

\subsection{Diffusion/Flow Matching Models for Single Table Synthesis} 
\label{subsec:DataAug_Single}

This subsection reviews works on \textit{diffusion and flow matching models for single table synthesis} (including over-sampling). For clarity, we subdivide related works into three sub-categories: 1) generic diffusion and flow matching models for tabular data (see Sec.~\ref{subsubsec:GenericSingleTableSynthesis}), 2) diffusion and flow matching models for tabular data in the healthcare domain (see Sec.~\ref{subsubsec:HealthcareTadDiff}), and 3) diffusion models for tabular data in the finance domain (see Sec.~\ref{subsubsec:FinanceTabDiff}). To ensure readers can follow along, we present the task-specific setup and problem definition below (see Problem~\ref{prob:SingleTableSynthesis}) and provide a summary of the models in Table~\ref{SummarySingleTableSynthesis}. Within each sub-category, we further separate diffusion-based and flow-matching-based methods when both are present.

Given a table $R$, we utilize $\prescript{j}{i}{\mathbf{x}_{t}}$ to denote the $j$-th feature value of the $i$-th sample at time point $t$.  On this basis, let the number of numerical features be  $M_{\text{num}}$ and the number of categorical features be $M_{\text{cat}}$. By reorganizing the order of the features, a sample (i.e., a row) can be represented as $ \mathbf{x} = (\mathbf{x}^{\text{num}}, \mathbf{x}^{\text{cat}}) $, where $ \mathbf{x}^{\text{num}} \in \mathbb{R}^{M_{\text{num}}} $ and $ \mathbf{x}^{\text{cat}} \in \mathbb{Z}^{M_{\text{cat}}} $. The $j$-th categorical feature can have $C_{j}$ distinct feature values $\{1,\cdots,C_{j}\}$, namely $\prescript{j}{}{\mathbf{x}^{\text{cat}}} \in \{1,\cdots,C_{j}\}$. With this, the \textit{single table synthesis problem} can be defined as:
\begin{problem}[Single Table Synthesis]\label{prob:SingleTableSynthesis}
 Given a table $R=\{\mathbf{x}\}$, we aim to learn a generative model $p_{\boldsymbol{\theta}}(R)$ to generate a synthetic table $\hat{R}=\{\hat{\textbf{x}}\}$ with high-quality and diversity.
\end{problem}

\begin{table}[h]
\centering
\caption{Overview of Diffusion and Flow Matching Models for Single Table Synthesis. `*' refers to a name created by us for convenience.}
\label{SummarySingleTableSynthesis}
\resizebox{\linewidth}{!}{%
\begin{tabular}{@{}llllll@{}}
\toprule
\textbf{Paradigm} & \textbf{Model Name} & \textbf{Year} & \textbf{Venue} & \textbf{Feature Type}  & \textbf{Domain}  \\ 
\midrule
\multirow{14}{*}{Diffusion} & \textbf{SOS} \cite{kim2022sos}  & 2022         & KDD            & Num by Default & Generic\\
& \textbf{STaSy} \cite{kimstasy}              & 2023         & ICLR           & Num+Cat & Generic      \\
& \textbf{TabDDPM} \cite{kotelnikov2023tabddpm}             & 2023         & ICML           & Num+Cat    & Generic   \\
&  \textbf{CoDi} \cite{lee2023codi}               & 2023         & ICML           & Num+Cat      & Generic \\
& \textbf{MissDiff} \cite{ouyang2023missdiff}               & 2023         & ICMLW           & Num+Cat  & Generic     \\
& \textbf{AutoDiff} \cite{suh2023autodiff}           & 2023         & NeurIPSW          & Num+Cat+Mixed & Generic \\
& \textbf{TabSyn} \cite{zhangmixed}              & 2024         & ICLR           & Num+Cat        & Generic \\
& \textbf{Forest-Diffusion} \cite{jolicoeur2024generating} & 2024 & AISTATS       & Num+Cat     & Generic   \\
& \textbf{TabDiff} \cite{shi2024tabdiff}            & 2025         & ICLR   & Num+Cat & Generic       \\
& \textbf{TabUnite} \cite{si2024tabunite}            & 2024         & OpenReview   & Num+Cat   & Generic    \\
& \textbf{CDTD} \cite{mueller2024continuous}            & 2025         & ICLR   & Num+Cat   & Generic    \\
& \textbf{TabRep} \cite{si2026tabrep}            & 2026         & TMLR   & Num+Cat   & Generic    \\
& \textbf{TabNAT} \cite{zhang2025tabnat}            & 2025         & ICML   & Num+Cat   & Generic    \\
& \textbf{TabuSDE} \cite{malashin2025stochastic}            & 2025         & ICDMW   & Num+Cat   & Generic    \\
\hline
\multirow{4}{*}{Flow Matching} & \textbf{TabbyFlow} \cite{guzman-cordero2025exponential}            & 2025         & ICML   & Num+Cat   & Generic    \\
& \textbf{TabSynFlow} \cite{nasution2026flow}            & 2026         & TMLR   & Num+Cat   & Generic    \\
& \textbf{TabFlowM} \cite{anonymous2026tabflowm}            & 2026         & TMLR Sub.   & Num+Cat   & Generic    \\
& \textbf{TabCascade} \cite{mueller2026cascaded}            & 2026         & ArXiv   & Num+Cat+Mixed   & Generic    \\
\hline
\multirow{6}{*}{Diffusion} & \textbf{MedDiff} \cite{he2023meddiff}  & 2023 & ArXiv & Num by default  & Healthcare\\
& \textbf{EHR-TabDDPM} \cite{ceritli2023synthesizing}  & 2023 & ArXiv & Num+Cat  & Healthcare\\
& \textbf{DPM-EHR}* \cite{nicholas2023synthetic}           & 2023         & NeurIPSW          & Num+Cat & Healthcare \\
& \textbf{FlexGen-EHR} \cite{he2024flexible} & 2024         & ICLR           & Cat+TS       & Healthcare \\
& \textbf{EHRDiff} \cite{yuan2024ehrdiff} & 2024         & TMLR           & Cat+Num+TS    & Healthcare  \\
& \textbf{EHR-D3PM} \cite{han2024guided}            & 2025         & TMLR   & Cat   & Healthcare    \\
\hline
Flow Matching & \textbf{PatientFlow} \cite{branco2026patientflow}            & 2026         & AI Med.   & Num+Cat+TS   & Healthcare    \\
\hline
\multirow{3}{*}{Diffusion} & \textbf{FinDiff} \cite{sattarov2023findiff}          & 2023         & ICAIF          & Num+Cat & Finance\\
& \textbf{EntTabDiff} \cite{liu2024entity}            & 2024         & ICAIF   & Num+Cat  & Finance      \\
& \textbf{Imb-FinDiff} \cite{schreyer2024imb}            & 2024         & ICAIF   & Num+Cat  & Finance      \\
\bottomrule
\end{tabular}%
}
\label{tab:tabular_data_synthesis}
\end{table}

\subsubsection{Generic Diffusion and Flow Matching Models for Single Table Synthesis}
\label{subsubsec:GenericSingleTableSynthesis}
We use ``generic'' to refer to methods intended as broadly applicable single-table generators, rather than models tied to a specific application domain. This does not imply that one model works uniformly well on every dataset; instead, these methods are formulated before adding assumptions such as EHR temporal structure or financial transaction rules. The main questions are therefore representational and procedural: how to encode numerical and categorical columns, whether to generate in data space or latent space, how to preserve dependencies and handle missing entries, and how flow matching changes probability paths and sampling cost. Following Table~\ref{SummarySingleTableSynthesis} and the taxonomy in Figure~\ref{fig:taxonomy}, we first discuss diffusion-based methods, then flow-matching-based methods, and finally compare their shared design choices and limitations.

\paragraph{Diffusion-based generic synthesis}

\textbf{SOS} \cite{kim2022sos} is the first to apply score-based generative models to tabular data oversampling. It trains class-specific score networks and generates minority-class samples either from random noise or from noised majority-class samples through a minority-class reverse process. This makes SOS mainly an oversampling method, but it shows that score-based generation can be useful for tabular data. Moving from oversampling to general data synthesis, \textbf{STaSy} \cite{kimstasy} proposes a score-based tabular synthesis method based on stochastic differential equations. It improves score learning with self-paced training and fine-tunes the pre-trained score network with a probability-flow ODE. However, STaSy still converts categorical features into one-hot representations and rounds them back after generation, which may introduce invalid categories and weaken dependency preservation.

Unlike STaSy, which follows a continuous diffusion framework, \textbf{TabDDPM} \cite{kotelnikov2023tabddpm} adapts diffusion models to mixed-type tabular data by modeling numerical and categorical features separately. It uses Gaussian diffusion for numerical features and multinomial diffusion for categorical features, which makes it a simple and influential baseline for generating mixed-type tables. However, because the two feature types are mostly modeled through separate corruption processes, cross-type dependency modeling remains a challenge. To better capture correlations between different feature types, \textbf{CoDi} \cite{lee2023codi} proposes a co-evolving diffusion framework with two coupled diffusion models: one for numerical attributes and the other for categorical attributes. Each reverse process is conditioned on the current noisy representation of the other feature type, and contrastive learning is used to align numerical and categorical representations. CoDi therefore improves dependency modeling over a purely separate design, but it still depends on preprocessing and postprocessing choices such as scaling, one-hot encoding, and rounding.

After TabDDPM and CoDi, a natural question is whether tabular diffusion models can better handle practical issues beyond the basic numerical/categorical split. \textbf{AutoDiff} \cite{suh2023autodiff} addresses \textit{heterogeneous features} by integrating an autoencoder with a score-based diffusion model. The autoencoder maps heterogeneous features into a continuous latent space, where diffusion is performed, and the decoder reconstructs samples in the original feature space. To handle \textit{mixed-type features}, AutoDiff introduces a dummy variable encoding the frequency of repeated values, and it learns \textit{feature correlations} through the joint distribution of latent embeddings. However, its performance and privacy depend strongly on the autoencoder: the authors observe weaker privacy guarantees, partly because a sophisticated autoencoder may overfit the input data and diffusion models may memorize training samples \cite{carlini2023extracting}.

Vanilla diffusion models typically require complete training data, while tabular data often suffers from missing values. Ouyang et al. \cite{ouyang2023missdiff} observe that an ``impute-then-generate'' pipeline may introduce learning bias because single imputation cannot capture data variability. To address this issue, \textbf{MissDiff} trains diffusion models directly on incomplete mixed-type tables without prior imputation. Its main idea is to mask the denoising score-matching loss over missing entries, while using simple placeholders for missing numerical and categorical values so that the score function remains well-defined. This makes MissDiff an early tabular diffusion model for learning from incomplete data, but it still follows common preprocessing and postprocessing choices such as min-max scaling, one-hot encoding, softmax, and rounding, so categorical validity and dependency preservation remain concerns.

Following the same motivation of avoiding direct mixed-type diffusion, \textbf{TabSyn} \cite{zhangmixed} explores latent diffusion for tabular data synthesis, although AutoDiff \cite{suh2023autodiff} addressed this idea earlier. The authors emphasize that simple encoding strategies, such as one-hot encoding and analog bit encoding, often lead to suboptimal performance, while employing separate models for different feature types hampers the ability to capture cross-feature correlations. To overcome these limitations, TabSyn uses a VAE with feature tokenizers to transform raw numerical and categorical features into continuous embeddings, trains a score-based diffusion model in the latent space, and then uses a detokenizer to recover real feature values from the generated embeddings. This joint latent space enables the model to handle different feature types while preserving their correlations. Ablation studies further show that replacing the VAE with simple one-hot encoding gives the poorest performance, indicating that treating categorical features as continuous without learned representations is inadequate. However, TabSyn cannot directly handle missing values during training; missing-value processing and feature transformations must be done in preprocessing, which may introduce bias or noise.

Rather than relying on neural networks as universal function approximators \cite{hornik1989multilayer} to estimate the score or noise function, \textbf{Forest-Diffusion} \cite{jolicoeur2024generating} employs XGBoost \cite{chen2016xgboost}, leveraging its strength in tabular data prediction and classification. The method duplicates the training data across discretized noise levels, adds noise or interpolates between data and noise, and trains XGBoost models to predict the score at different time points. An important advantage is that XGBoost can naturally handle missing data by learning the best split, allowing Forest-Diffusion to be trained directly on incomplete datasets. For categorical data, it still uses dummy encoding during preprocessing and rounds dummy variables to the nearest class during postprocessing. Forest-Diffusion can also be adapted to data imputation and conditional synthesis, but its main message is more general: tabular diffusion models do not have to rely only on neural denoisers, and strong tabular learners can be inserted into the diffusion framework. At the same time, categorical postprocessing and the need to train models over many noise levels remain practical concerns.

After latent-space and tree-based alternatives, another line of work revisits how mixed features should be represented and noised within diffusion itself. \textbf{TabDiff} \cite{shi2024tabdiff} identifies several limitations in prior methods, including the encoding overhead of latent diffusion models, the imperfect handling of discrete-time diffusion in mixed-type tables, and the insufficient treatment of feature-wise distribution heterogeneity. To address these issues, TabDiff adopts a continuous-time diffusion framework with learnable noise schedules for individual features. Numerical features are modeled with an SDE-based diffusion process, while categorical features are handled with masking diffusion. Although the paper describes this as a joint diffusion process, the numerical and categorical parts are still modeled differently; its main contribution is therefore feature-adaptive noise scheduling rather than a fully unified mixed-type diffusion process. \textbf{TabUnite} \cite{si2024tabunite} pushes the representation question further. It argues that separate generative processes and simple encodings such as one-hot representations can weaken cross-feature correlation modeling, while learned latent embeddings may be parameter inefficient. It therefore uses a Quantile Transformer for continuous features and compact encodings such as analog bits, PSK encoding, or dictionary encoding for categorical features, then applies a single diffusion or flow model in the unified data space. Together, these works show that mixed-type tabular generation depends not only on the denoising model, but also on how feature types are encoded and how noise schedules are assigned across heterogeneous columns.

Along the same line, \textbf{CDTD} \cite{mueller2024continuous} points out that existing diffusion models for numerical and categorical features are mostly built upon advances from image or text domains, so their noise schedules are not specifically designed for mixed tabular features. This may lead to unbalanced losses across feature types, poor scalability, and weak performance on categorical features with many categories. CDTD addresses these challenges by embedding categorical features into a continuous space and applying a unified Gaussian diffusion process. It further explores learnable noise schedules at different granularities, including a single schedule for all features, separate schedules for numerical and categorical features, and individual schedules per feature. This design improves mixed-feature handling and high-cardinality categorical modeling, but it also shows that schedule design and loss balancing remain central issues for tabular diffusion.

\textbf{TabRep} \cite{si2026tabrep} further emphasizes that representation design is not merely a preprocessing detail. Instead of using sparse one-hot encodings or an additional latent model, TabRep introduces CatConverter, which maps each categorical value to a two-dimensional cosine-sine representation on a unit circle and concatenates it with quantile-transformed numerical features. A DDPM or flow matching model can then be trained in this unified continuous space, and generated categorical values are recovered by projecting the output to the nearest valid category. This compact representation avoids some drawbacks of sparse categorical encodings, but it also introduces a new assumption: categorical values must be ordered before placement on the unit circle. Different orderings can affect performance, suggesting that future tabular diffusion models may need semantically informed or learnable categorical representations.

Recent work also tries to combine diffusion with more structured dependency modeling. \textbf{TabNAT} \cite{zhang2025tabnat} approaches mixed-type tabular synthesis by combining masked generative modeling with diffusion. Its motivation is that autoregressive models can explicitly model column dependencies but struggle with continuous values and fixed generation orders, while diffusion models handle continuous variables naturally but may underuse column dependency structure. TabNAT embeds continuous features and discrete columns into a shared sequence, randomly masks target positions, and uses a bidirectional Transformer to produce conditional embeddings. Discrete columns are generated with categorical prediction, while continuous columns are generated with a conditional diffusion model. This design supports order-agnostic generation and can also be used for missing-value imputation, although its imputation setting assumes complete training data and evaluates artificially injected missing entries.

\textbf{TabuSDE} \cite{malashin2025stochastic} takes a more system-level route for joint tabular synthesis and completion. It uses learnable per-feature drift and diffusion coefficients to adapt SDEs to heterogeneous columns, combines categorical embeddings with transformations for numerical variables, and uses cross-feature attention to capture inter-column dependencies. It also predicts feature values and missingness masks, so that structured missing-data patterns can be modeled during generation. The framework further includes privacy accounting and accelerated sampling, but these many components make the source of performance gains harder to isolate. Therefore, TabuSDE is best viewed as a broad design that combines feature-specific SDEs, missingness modeling, attention, privacy accounting, and sampling acceleration, rather than as evidence that any single component alone solves tabular synthesis and completion.

\paragraph{Flow-matching-based generic synthesis}

Flow matching extends this discussion by replacing denoising-based generation with learned transport paths. \textbf{TabbyFlow} \cite{guzman-cordero2025exponential} is a dedicated flow-matching framework for mixed-type tabular synthesis. Instead of forcing all columns into a single Gaussian score-matching objective, it uses exponential-family variational posteriors so that numerical variables and categorical variables can be handled with feature-type-specific losses under a common variational flow matching view. This makes TabbyFlow important not only as a new generator, but also as an example of how flow matching can be adapted to heterogeneous tabular likelihoods.

Beyond direct data-space flow matching, another question is how flow matching should be used when the representation or feature type is more specialized. \textbf{TabFlowM} \cite{anonymous2026tabflowm} studies whether a lightweight latent-space flow-matching generator can replace diffusion-style score learning after mixed-type records are mapped into a decoder-compatible continuous token space. It trains a time-conditioned velocity field on a simple coupling path and samples by integrating the learned ODE. This makes TabFlowM useful evidence that latent flow matching can reduce training or sampling cost, while also showing that performance still depends on the quality of the learned latent representation.

The same representation issue becomes sharper when numerical columns contain discrete states rather than ordinary continuous values. In many tables, missing entries, censored observations, or zero-inflated outcomes behave like special states, while the remaining values still require continuous modeling. \textbf{TabCascade} \cite{mueller2026cascaded} addresses this setting with a coarse-to-fine flow-matching pipeline: it first generates categorical features and discretized numerical states at low resolution, and then conditions a high-resolution flow model on this coarse output to recover numerical details. This design is useful for difficult mixed-type features, although its missingness setting is simulated and it does not provide formal privacy guarantees.

A related question is how much of flow matching's behavior comes from the transport formulation itself, and how much comes from the representation on which it is applied. \textbf{TabSynFlow} \cite{nasution2026flow} studies this question by replacing TabSyn's latent diffusion model with a flow-matching velocity field, and comparing this latent-space design with data-space variants such as TabbyFlow under different paths and samplers. Its results suggest that flow matching can often reach competitive utility with fewer function evaluations, but privacy risk remains a post-hoc quantity that depends on the dataset, representation, and path choice.

\paragraph{Comparative Analysis} Overall, generic single-table methods are best compared along three design axes. The first is representation. STaSy treats categorical variables through one-hot encodings in a continuous score/SDE framework, while TabDDPM and CoDi separate numerical and categorical corruption channels. AutoDiff, TabSyn, TabUnite, and TabRep instead map heterogeneous columns into latent or unified continuous spaces before applying diffusion or flow matching. This makes the generative dynamics easier to define, but shifts part of the burden to encoding and decoding: category validity, rare values, and semantic distances depend on the representation. The second axis is where tabular heterogeneity is handled. TabSyn and TabRep mainly adapt the feature space to standard diffusion or flow matching, whereas TabDiff, CDTD, and TabuSDE adapt the process itself through schedules, losses, masks, or feature-wise dynamics. This process-level direction is useful because tabular columns have unequal scales, cardinalities, missingness patterns, and marginal distributions. However, it also adds hyperparameters and makes attribution harder: when performance improves, the gain may come from the schedule, backbone, missingness model, representation, or tuning budget. The third axis is path design and efficiency. Flow matching expands the choices of probability path and sampler: TabbyFlow uses feature-type-aware variational objectives for mixed variables, while TabFlowM and TabCascade use latent or cascaded transport to reduce sampling cost or handle difficult mixed-type features. These results suggest that faster generation is possible, but efficiency is not a proxy for quality or safety. Conclusions still depend on downstream utility, fidelity, disclosure risk, rare-event behavior, and tuning protocol. Thus, the practical question is not simply whether diffusion or flow matching models can synthesize tables, but which representation, generative process, and evaluation protocol fit the target application.

\subsubsection{Diffusion and Flow Matching Models for Single Table Synthesis in Healthcare Domain} 
\label{subsubsec:HealthcareTadDiff}
Here, we review diffusion and flow matching models for single-table synthesis in the healthcare domain, with a primary focus on electronic health records (EHRs). EHR data often combines demographics, diagnoses, medications, laboratory measurements, and treatment outcomes, making it valuable for medical research but difficult to share directly because of privacy risks~\cite{yadav2018mining}. Synthetic EHR generation therefore aims to preserve useful clinical patterns while reducing the need to release patient records~\cite{han2024guided}.

\paragraph{Diffusion-based healthcare synthesis}

Early diffusion-based EHR generators mainly adapt general tabular diffusion ideas to clinical data. \textbf{MedDiff} \cite{he2023meddiff} uses a DDIM-style generator with accelerated sampling and classifier guidance for conditional generation, but it is limited to continuous features. \textbf{EHR-TabDDPM} \cite{ceritli2023synthesizing} instead applies TabDDPM \cite{kotelnikov2023tabddpm} to mixed-type EHR tables, making it a more direct baseline for numerical and categorical clinical attributes. Both works evaluate synthetic quality through fidelity and downstream utility, with EHR-TabDDPM also considering privacy.

After early static EHR generators, later work turns to longitudinal and heterogeneous clinical records. \textbf{DPM-EHR} \cite{nicholas2023synthetic} extends diffusion-based synthesis to longitudinal mixed-type EHRs, using one-hot categorical representations together with Gaussian diffusion. This makes it an early attempt to move beyond static EHR tables, although discrete medical codes are still treated through a continuous representation. \textbf{FlexGen-EHR} \cite{he2024flexible} further considers a more realistic setting where static and temporal modalities may be missing not completely at random. It encodes temporal and static features separately, aligns their latent spaces with an optimal-transport module, applies latent diffusion, and decodes both modalities back to EHR records. This design is better matched to heterogeneous hospital data, but the added encoders, alignment module, and decoders also make it harder to isolate which component drives the improvement.

Another design choice is whether medical codes should be converted into a continuous space or modeled as discrete variables. \textbf{EHRDiff} \cite{yuan2024ehrdiff} uses a real-valued representation for both categorical and continuous EHR features and applies an SDE-based diffusion model with probability-flow ODE sampling. This gives a simple unconditional generator, but the meaning of categorical codes still depends on the chosen encoding and normalization. \textbf{EHR-D3PM} \cite{han2024guided} takes the opposite route by modeling categorical medical codes directly with a D3PM-style discrete diffusion process \cite{austin2021structured} and adding training-free classifier/energy guidance for conditional generation. This is better aligned with code-based EHR data, although continuous measurements and richer temporal structures still require additional modeling.

\paragraph{Flow-matching-based healthcare synthesis}

Flow matching in healthcare is represented by \textbf{PatientFlow} \cite{branco2026patientflow}, which moves beyond static EHR tables to mixed-type longitudinal clinical data. It learns latent representations for static patient information and temporal assessments, then uses a static latent flow and a temporal flow conditioned on the generated static representation to synthesize patient trajectories. Its evaluation is broader than generic fidelity, covering longitudinal similarity, train-on-synthetic/test-on-real prognosis, clinical rules, expert discrimination, and privacy risk. At the same time, reported violations of onset and ALSFRS-R progression rules show that longitudinal clinical synthesis still needs domain-semantic validation, not just distributional similarity or downstream utility.

\paragraph{Comparative Analysis}
Healthcare synthesis is constrained by clinical meaning rather than table format alone. MedDiff and EHR-TabDDPM show how general tabular diffusion can be adapted to EHR records, whereas DPM-EHR, FlexGen-EHR, and PatientFlow move toward longitudinal or heterogeneous patient trajectories. EHRDiff and EHR-D3PM highlight a complementary representation issue: continuous encodings are convenient, but discrete diagnosis or procedure codes may require discrete generative processes. Evaluation therefore needs to go beyond distributional fidelity and downstream utility by also checking privacy risk, clinical rules, and temporal consistency. Otherwise, a synthetic record may look statistically plausible while violating medical semantics.

\subsubsection{Diffusion Models for Single Table Synthesis in Finance Domain}
\label{subsubsec:FinanceTabDiff}
In finance, diffusion models are mainly used to adapt mixed-type tabular synthesis to sensitive transactions, entity identifiers, and class imbalance, which are central to risk modeling, fraud detection, and privacy-preserving data sharing \cite{assefa2020generating}. \textbf{FinDiff} \cite{sattarov2023findiff} targets mixed-type financial tables by replacing categorical values with continuous embeddings, concatenating them with normalized numerical features and conditioning embeddings, and training a DDPM in the resulting representation space. The generated embeddings are then mapped back to valid categories by nearest-neighbor decoding. This design makes diffusion easier to apply to financial tables with categorical attributes, but its quality depends on whether the learned embeddings preserve useful category semantics.

However, representing each categorical column with embeddings is still not enough when a column denotes an entity, such as a company or an individual. \textbf{EntTabDiff} \cite{liu2024entity} therefore treats entity columns as a special part of the financial table rather than as ordinary categories. Instead of only reproducing records for entities observed in the training data, it learns an entity distribution and generates tabular records conditioned on synthetic entities through cross-attention. This makes the model more suitable for entity-aware financial synthesis, although the generated records still depend on how well the entity distribution and entity-feature relationships are learned.

A second finance-specific issue is that the most important classes are often rare, as in fraud or high-risk detection. \textbf{Imb-FinDiff} \cite{schreyer2024imb} extends FinDiff in this direction by adding class conditioning and an auxiliary class-label prediction objective, encouraging the diffusion model to generate minority-class records more effectively. Taken together, these finance-oriented methods show that the main issue is not a new diffusion backbone, but how to inject financial structure, including categorical semantics, entity identity, and class imbalance, into representation and conditioning design.

\subsection{Diffusion Models for Multi-relational Data Synthesis}
\label{subsec:DataAug_Multi}

Most tabular data synthesis methods focus on single-table generation (see Sec.~\ref{subsec:DataAug_Single}); however, real-world datasets often consist of multiple interconnected tables, highlighting the importance of multi-relational data synthesis---a challenge that remains largely underexplored. In this subsection, we first present the task-specific setup and problem definition (see Problem~\ref{prob:MultiRelationalSynthesis}). Then, we review related works in chronological order, with a summary provided in Table~\ref{SummaryMultiTableSynthesis}.

In addition to capturing the characteristics of individual tables, multi-relational data synthesis requires modeling relationships across multiple tables that are connected (i.e., constrained by foreign keys). A multi-relational database $\mathcal{R}$ consists of $P$ tables $\{R_{1},...,R_{P}\}$. Each table $R_{p}$ has \textit{primary key} which is the unique identifiers of rows. Given two tables  $R_{p}$ and  $R_{q}$, we say $R_{p}$ refers to $R_{q}$ (or $R_{p}$ has a \textit{foreign key constraint} with $R_{q}$, or $R_{p}$ has a \textit{parent-child relationship} with $R_{q}$ where $R_{q}$ is the \textit{parent} and $R_{p}$ is the \textit{child}) if $R_{p}$ has a feature (called \textit{foreign key}) that refers to the primary key of $R_{q}$. With these notations, the \textit{multi-relational data synthesis problem} can be defined as follows \cite{pang2024clavaddpm}:
\begin{problem}[Multi-relational Data  Synthesis]\label{prob:MultiRelationalSynthesis}
   Given $\mathcal{R} = \{R_{1},...,R_{P}\}$, we aim to generate a synthetic database $\tilde{\mathcal{R}} = \{\tilde{R}_{1},...,\tilde{R}_{P}\}$ that preserves the structure, foreign-key constraints, as well as correlations among features, including any inter-column correlations within the same table, inter-table correlations, and intra-group correlations.
\end{problem}

\begin{table}[h]
\centering
\caption{Overview of Diffusion Models for Multi-relational Data Synthesis}
\label{SummaryMultiTableSynthesis}
\resizebox{\columnwidth}{!}{
\begin{tabular}{@{}llllll@{}}
\toprule
\textbf{Paradigm} & \textbf{Model Name} & \textbf{Year} & \textbf{Venue} & \textbf{Feature Type} & \textbf{Domain}\\ 
\midrule
\multirow{2}{*}{Diffusion} & \textbf{ClavaDDPM} \cite{pang2024clavaddpm}             & 2024         & NeurIPS           & Num by default  & Generic     \\
& \textbf{GNN-TabSyn} \cite{hudovernik2024relational}  & 2024         & NeurIPSW            & Num+Cat & Generic \\
\bottomrule
\end{tabular}%
}
\end{table}

Diffusion-based multi-relational synthesis still builds on single-table generators, but the key additional challenge is how to preserve dependencies across tables connected by foreign keys. \textbf{ClavaDDPM} \cite{pang2024clavaddpm} addresses this by introducing clustering labels as relational intermediaries: instead of synthesizing each table independently, it uses these labels to summarize cross-table dependencies and then applies a TabDDPM-style generator \cite{kotelnikov2023tabddpm} in a unified continuous feature space. This design is relatively simple and scalable, and it can capture group-level dependencies, but its success depends on whether the clustering labels preserve the relational information that matters for downstream queries.

\textbf{GNN-TabSyn} \cite{hudovernik2024relational} takes a more structural view. It represents a relational database as a heterogeneous graph, where rows are nodes and foreign-key links are edges, and then uses GNN embeddings as conditions for a latent TabSyn generator \cite{zhangmixed}. This makes inter-table topology explicit and is better aligned with the relational nature of the data. However, the method also inherits the cost and assumptions of graph representation learning, and it does not naturally synthesize unseen graph structures. Overall, these two methods show that multi-relational synthesis is less about generating realistic rows in isolation, and more about preserving join behavior, foreign-key structure, and cross-table dependencies.

Compared with single-table synthesis, this area remains underdeveloped: there are only a few diffusion-based systems, no mature flow-matching counterpart, and limited agreement on benchmarks or metrics. A promising direction is to combine transport-based generation with explicit relational constraints, so that probability paths or noising processes operate over both row attributes and relational structure. However, such methods would need to handle schema validity, unseen keys, rare relationship patterns, and privacy leakage caused by linkability across tables.

\section{Diffusion Models for Data Imputation}
\label{sec:DataImp}

Missing values imputation is a long-standing research problem in data mining and machine learning community \cite{lin2020missing}. These approaches can be divided into two categories \cite{jarrett2022hyperimpute}: 1) \textit{iterative approaches} that estimate the conditional distribution of one feature based on other features (e.g., MICE\cite{van2000multivariate}); and 2) \textit{deep generative approaches} that train a generative model to generate values in missing parts based on observed parts (e.g., MIDA \cite{gondara2018mida} and MIWAE \cite{mattei2019miwae} based on denoising AEs, HIVAE \cite{nazabal2020handling} based on VAEs, and GAIN \cite{yoon2018gain} based on GANs). We formally define the missing data imputation problem below (see Problem~\ref{prob:TabularDataImputation}) and summarize representative diffusion-based imputation methods in Table~\ref{SummaryDataImputation}. We focus on diffusion here because dedicated flow-matching methods for tabular imputation have not yet formed a comparable literature; existing tabular flow-matching work mainly studies synthesis or anomaly scoring rather than missing-value recovery.

\begin{problem} [Tabular Data Imputation]\label{prob:TabularDataImputation}
Given a $D$-dimensional training dataset $\mathbf{X} = \{\prescript{}{i}{\mathbf{x}}\}_{i=1}^{N}$, a feature $j \in \{1,...,D\}$ of $\prescript{}{i}{\mathbf{x}}$ is denoted as $\prescript{j}{i}{\mathbf{x}}$, where the $j$-th feature can be numerical or categorical. Besides, any feature $j$ can suffer from missing values. Let $\mathcal{X} = (\mathbb{R}\cup\emptyset)^{D}$ be the input space, where $\mathbb{R}$ denotes the real number space for a numerical feature. For brevity, we also utilize $\mathbb{R}$ to denote the corresponding range if it is  a categorical feature. Data imputation aims to find a function $f(\cdot)$: $\mathcal{X} \to \mathbb{R}^{D}$ which can replace the missing values with reasonable values.
\end{problem}
%Maybe we can replace the problem statement with another one in \cite{chen2024rethinking}

\begin{table}[h]
\centering
\caption{Overview of Diffusion Models for Data Imputation}
\label{SummaryDataImputation}
\resizebox{\columnwidth}{!}{
\begin{tabular}{@{}llllll@{}}
\toprule
\textbf{Paradigm} & \textbf{Model Name} & \textbf{Year} & \textbf{Venue} & \textbf{Feature Type}  & \textbf{Domain} \\ 
\midrule
\multirow{10}{*}{Diffusion} & \textbf{TabCSDI} \cite{zheng2022diffusion}  & 2022         & NeurIPSW            & Num+Cat & Generic\\
& \textbf{TabDiff} \cite{shi2024tabdiff}            & 2025         & ICLR   & Num+Cat     & Generic   \\
& \textbf{SimpDM} \cite{liu2024self}            & 2024         & CIKM   & Num+Cat    & Generic    \\
& \textbf{MTabGen} \cite{villaizan2024diffusion}            & 2025         & TKDD   & Num+Cat     & Generic  \\
& \textbf{DDPM-Perlin} \cite{wibisono2024natural}            & 2024         & KBS   & Num   & Generic    \\
& \textbf{NewImp} \cite{chen2024rethinking}            & 2024         & NeurIPS   & Num  & Generic    \\
& \textbf{DiffPuter} \cite{zhang2024unleashing}            & 2025         & ICLR   & Num+Cat  & Generic   \\
& \textbf{HARPOON} \cite{shankar2026harpoon}            & 2026         & ICLR   & Num+Cat  & Generic   \\
& \textbf{TabNAT} \cite{zhang2025tabnat}            & 2025         & ICML   & Num+Cat  & Generic   \\
& \textbf{TabuSDE} \cite{malashin2025stochastic}            & 2025         & ICDMW   & Num+Cat  & Generic   \\
\bottomrule
\end{tabular}%
}
\end{table}

Early diffusion-based imputation methods formulate missing-value recovery as conditional denoising. \textbf{TabCSDI} \cite{zheng2022diffusion} extends CSDI \cite{tashiro2021csdi} from time-series imputation to tabular data by splitting a record into observed and target missing entries, and then training a conditional diffusion model to denoise the target part given the observed part. To support mixed-type tables, categorical features are converted by one-hot encoding, analog bits \cite{chenanalog}, or feature tokenization \cite{gorishniy2021revisiting}, and the imputed outputs are decoded back to the original feature space. This provides a simple template for diffusion-based imputation, but categorical validity and dependency preservation still depend on the encoding and decoding choices.

\textbf{TabDiff} \cite{shi2024tabdiff}, originally proposed for data synthesis (see Sec.~\ref{subsec:DataAug_Single}), follows a similar conditional-denoising view for imputation. Observed features are fixed during the reverse process, while missing numerical and categorical features are generated with classifier-free guidance. Compared with TabCSDI, TabDiff benefits from a diffusion design already tailored to mixed-type tabular data, but its imputation setting is still mainly an adaptation of a synthesis model rather than a method designed specifically for incomplete-data learning.

The next group of methods tries to make conditional diffusion more suitable for the accuracy-driven nature of imputation. \textbf{SimpDM} \cite{liu2024self} starts from the observation that generation and imputation have different goals: diffusion models are designed to produce diverse samples from noise, whereas imputation usually requires stable and accurate values for a fixed observed context. It therefore adds a self-supervised alignment mechanism so that imputations from the same observed entries remain consistent, and uses perturbation-based data augmentation to reduce the effect of small tabular training sets. For mixed-type features, however, it still follows the TabDDPM-style separation of Gaussian diffusion for numerical features and multinomial diffusion for categorical features \cite{kotelnikov2023tabddpm}, so it inherits similar mixed-type modeling limitations.

\textbf{MTabGen} \cite{villaizan2024diffusion} takes a broader route by turning imputation and synthesis into two cases of the same masked-generation framework. It normalizes numerical features, ordinally encodes categorical features, projects heterogeneous columns into a continuous latent space, and uses a Transformer encoder-decoder with attention-based conditioning to model the relation between observed and missing features. Dynamic masking allows the model to handle different visible-feature patterns; when all features are masked, the same framework becomes data synthesis. This makes MTabGen more flexible than a pure imputation model, but the unified latent representation and ordinal categorical encoding still make feature semantics and decoding quality important.

After conditional-denoising methods, another line asks whether the standard diffusion process itself is suited to imputation. \textbf{DDPM-Perlin} \cite{wibisono2024natural} follows the TabCSDI-style observed/target split, but replaces Gaussian corruption with Perlin noise when imputing numerical features. This explores whether structured noise can be more natural for imputation than independent Gaussian perturbations. However, the replacement is not well supported by the standard DDPM derivation, since Perlin noise does not have the same additive Gaussian properties.

\textbf{NewImp} \cite{chen2024rethinking} revisits the objective side of the same problem. It argues that diffusion models may generate unnecessarily diverse samples for a task that often needs accurate point estimates, and that mask-based training can be mismatched with the missingness mechanism at test time. Under a Wasserstein-gradient-flow view \cite{mokrov2021large}, NewImp introduces a cost functional with negative entropy regularization to reduce sample diversity and uses a joint-distribution formulation to avoid relying on a masking-based conditional objective. This gives a more imputation-oriented formulation, but it currently handles only numerical features.

A related issue is whether diffusion imputation should be treated as ordinary conditional sampling or as an incomplete-data optimization problem. \textbf{DiffPuter} \cite{zhang2024unleashing} argues that predictive imputers often outperform generative imputers because estimating a joint distribution is difficult when missing entries are unknown. It addresses this by integrating an EM-style procedure \cite{dempster1977maximum} with diffusion: the model alternates between learning a joint distribution over observed entries and current missing-value estimates, and updating the missing entries through reverse denoising. This makes diffusion imputation closer to incomplete-data learning, but mixed-type handling still relies on one-hot encoding followed by Gaussian diffusion over the encoded features.

\textbf{HARPOON} \cite{shankar2026harpoon} extends inference-time conditioning beyond missing-value imputation. It trains one unconditional data-space DDPM on standardized numerical and one-hot categorical features, and then guides sampling with differentiable constraints. The denoiser acts as a projection toward the learned data manifold, while constraint gradients move samples along this manifold, so missing-value constraints, ranges, categorical conditions, conjunctions, and disjunctions can be handled without retraining. Experiments on eight datasets show strong imputation performance and much lower violation rates than DiffPuter and GReaT under inequality constraints. However, HARPOON still works in a continuous encoded space and trains after filtering incomplete rows; it is therefore best viewed as a reusable constraint-guided sampler rather than a model that learns directly from naturally incomplete tables.

\textbf{TabNAT} \cite{zhang2025tabnat} represents a different setting. It treats imputation as flexible conditional generation from a masked bidirectional Transformer trained on complete data. Missing target columns are sampled in a random order conditioned on observed columns; discrete targets are sampled from predicted categorical distributions, while continuous targets use a RePaint-style denoising procedure. This is useful for heterogeneous conditional inference, but its imputation experiment injects MCAR missingness into the test set after complete-data training, so it should not be read as a method that learns directly from incomplete training tables.

\textbf{TabuSDE} \cite{malashin2025stochastic} moves one step further by modeling the missingness pattern together with the feature values. During conditional reverse diffusion, observed entries are clamped and only missing entries are updated. Its denoiser predicts both feature values and missingness probabilities, and a VAE-learned missingness prior helps model structured MAR/MNAR masks rather than independent random masks. This makes TabuSDE important because it treats missingness as part of the data-generating process, not merely as an inference-time mask. At the same time, it is a large integrated system, combining feature-specific SDEs, hybrid encoders, cross-feature attention, mask modeling, privacy accounting, hyperparameter search, and accelerated sampling, so its results support the promise of joint value--mask diffusion but do not isolate a single mechanism as the universal solution for tabular imputation.

\textbf{Comparative Analysis}: Taken together, the methods in Table~\ref{SummaryDataImputation} turn diffusion into an imputer in several related ways. One line treats imputation as conditional generation from a joint model trained on complete data, as in TabDiff, MTabGen, and TabNAT; this is suitable when a clean reference table is available, but it mainly evaluates test-time masking rather than learning from incomplete data. A second line changes the objective or sampling process to better match incomplete observations, as in TabCSDI, DDPM-Perlin, NewImp, and DiffPuter; these methods are closer to observational databases, but remain sensitive to the missingness mechanism, feature encoding, and observed-entry clamping. HARPOON adds an inference-time guidance view: missing entries become one type of condition, so the same sampler can also enforce range, categorical, conjunctive, or disjunctive constraints, although it still trains on complete rows. TabuSDE adds a missingness-aware direction by modeling values and masks jointly, which is important when missingness itself carries signal.

These distinctions make the evaluation protocol as important as the model design. MCAR masking on complete test data, MAR/MNAR missingness during training, and naturally incomplete benchmark tables test different abilities. Entry-wise RMSE, categorical accuracy, downstream utility, distributional distance, correlation preservation, and uncertainty calibration also capture different failure modes. A useful imputation study should therefore report how missingness is created or observed, whether the training table is complete, how observed entries are clamped during sampling, and whether multiple-imputation uncertainty is preserved. From this perspective, conditional denoising is reasonable for complete-training and simple test-time missingness, while mask-aware, incomplete-data, or value--mask objectives are better aligned with structured or informative missingness. Flow matching could in principle support similar conditional generation, but current tabular flow-matching studies have not yet established imputation-specific paths, mask handling, or uncertainty protocols.

\section{Diffusion Models for Trustworthy Data Synthesis}
\label{sec:TrustSynth}

Trustworthy data synthesis asks whether synthetic tables can be released and used without exposing individuals, reinforcing group bias, or reproducing training records. We begin with privacy-preserving synthesis. Two deployment settings are formalized below: cross-silo tabular synthesis, where different parties hold different feature subsets of the same records and want to synthesize a complete table without exposing raw feature values (Problem~\ref{prob:CrossSiloTabularSynthesis}), and federated tabular synthesis, where clients hold different records and collaboratively train a generator without centralizing local data (Problem~\ref{prob:FederatedTabularSynthesis}). We also discuss centralized differential privacy, where data are held by one curator but each record's influence on the learned generator must be bounded. We then review fairness-preserving synthesis (Problem~\ref{prob:FairnessPreservingSynthesis}) and memorization auditing. The reviewed methods are diffusion-based because dedicated flow-matching methods for privacy-preserving, fairness-aware, or memorization-aware tabular synthesis have not yet formed a comparable literature. Recent tabular flow-matching generators may report empirical disclosure-risk metrics in generic synthesis settings, but they do not yet provide explicit threat models, formal privacy mechanisms, fairness interventions, or memorization-mitigation protocols for trustworthy data release. We summarize representative methods in Table~\ref{SummaryTrustDataSythesis}.
\begin{problem}[Cross-Silo Tabular Synthesis]\label{prob:CrossSiloTabularSynthesis}
  Consider there are $L$ distinct parties $\{P_{1},...,P_{L}\}$, each party stores a subset of features $\mathbf{X}_{l} \in \mathbb{R}^{N\times D_{l}}$ with $D_{l}$ the number of features stored at party $P_{l}$. Altogether, we have $\mathbf{X}=\mathbf{X}_{1}||\mathbf{X}_{2}||...||\mathbf{X}_{L}$ (``$||$" means column-wise concatenation) and $\mathbf{X} \in \mathbb{R}^{N\times D}$ with $D=\sum_{l=1}^{L}D_{l}$. The goal of cross-silo tabular synthesis is to generate a synthetic dataset $\tilde{\mathbf{X}}=\tilde{\mathbf{X}}_{1}||\tilde{\mathbf{X}}_{2}||...||\tilde{\mathbf{X}}_{L}$ that are distributionally similar to the original dataset $\mathbf{X}$ while maintaining the private information of the actual values.  
\end{problem}

\begin{problem}[Federated Tabular Synthesis]\label{prob:FederatedTabularSynthesis} Consider there are $M$ distinct clients $\{Q_{1},...,Q_{M}\}$, each client stores a subset of samples $\mathbf{X}_{m} \in \mathbb{R}^{N_{m}\times D}$ with $N_{m}$ the number of samples stored at client $Q_{m}$. Altogether, we have $\mathbf{X} = \mathbf{X}_{1} \oplus \mathbf{X}_{2} \oplus \dots \oplus \mathbf{X}_{M}$ (``$\oplus$" denotes row-wise concatenation), and $\mathbf{X} \in \mathbb{R}^{N\times D}$ with with $N=\sum_{m=1}^{M}N_{m}$. The goal of federated learning based tabular synthesis is to generate a synthetic dataset $\tilde{\mathbf{X}}=\tilde{\mathbf{X}}_{1}\oplus\tilde{\mathbf{X}}_{2}\oplus...\oplus\tilde{\mathbf{X}}_{M}$ that are distributionally similar to the original dataset $\mathbf{X}$ while maintaining the private information of the actual values.  
\end{problem}

Meanwhile, \textit{fairness-preserving data synthesis} aims to use diffusion models to generate synthetic tabular data that maintains fairness w.r.t. sensitive features (e.g., age, gender, race, etc.) while preserving data utility. Formally, it is defined as:

\begin{problem} [Fairness-Preserving Data Synthesis]\label{prob:FairnessPreservingSynthesis}
Given a $D$-dimensional dataset $\mathbf{X} = \{\prescript{}{i}{\mathbf{x}}\}_{i=1}^{N}$, where each data point $\prescript{}{i}{\mathbf{x}} \in \mathbb{R}^{D}$ represents a record with $D$ features, let $\mathbf{S} \subseteq \{1, \ldots, D\}$ denote the set of indices corresponding to sensitive features. The goal is to learn a generative function $g(\cdot)$ to generate synthetic data $\hat{\mathbf{X}} = g(\mathbf{X})$ such that: 1) The synthetic data $\hat{\mathbf{X}} = \{\prescript{}{i}{\hat{\mathbf{x}}}\}_{i=1}^{N}$ preserves fairness with respect to sensitive features in $\mathbf{S}$, ensuring no unfair bias is introduced; and 2) The synthetic data $\hat{\mathbf{X}}$ maintains the statistical properties and utility of the original dataset $\mathbf{X}$ for downstream tasks.
\end{problem}

\begin{table}[h]
\centering
\caption{Overview of Diffusion Models for Trustworthy Data Synthesis.`*' refers to a name created by us for convenience.}
\label{SummaryTrustDataSythesis}
\resizebox{\columnwidth}{!}{%
\begin{tabular}{@{}llllll@{}}
\toprule
\textbf{Paradigm} & \textbf{Model Name} & \textbf{Year} & \textbf{Venue} & \textbf{Feature Type} & \textbf{Domain} \\ 
\midrule
\multirow{7}{*}{Diffusion} & \textbf{SiloFuse} \cite{shankar2024silofuse}  & 2024         & ICDE & Num+Cat  & Generic \\
& \textbf{FedTabDiff} \cite{sattarov2024fedtabdiff}  & 2024         & ArXiv   & Num+Cat  & Generic   \\
& \textbf{FairTabDDPM}* \cite{yang2024balanced}            & 2025         & TMLR   & Num+Cat  & Generic    \\
& \textbf{DP-Fed-FinDiff} \cite{sattarov2024differentially}            & 2024         & ArXiv   & Num+Cat  & Finance    \\
& \textbf{DP-FinDiff} \cite{sattarov2025privacy}            & 2025         & ArXiv   & Num+Cat  & Generic    \\
& \textbf{TabCutMix/TabCutMixPlus} \cite{fang2025understanding}            & 2025         & ICML   & Num+Cat  & Generic    \\
& \textbf{DynamicCut} \cite{fang2025closer}            & 2026         & TMLR   & Num+Cat  & Generic    \\
\bottomrule
\end{tabular}%
}
\end{table}
\textit{Cross-silo tabular synthesis} is the first privacy-preserving setting. \textbf{SiloFuse} \cite{shankar2024silofuse} targets vertically partitioned tables, where different silos store different feature subsets of the same records. It uses local autoencoders to map silo-side features, including categorical features, into continuous embeddings, and then trains a central latent DDPM over the aggregated embeddings. This allows the generator to model cross-silo feature correlations without directly sharing raw feature values. The design is especially useful for sparse or high-cardinality tables, although TabDDPM-style Gaussian/multinomial diffusion can still be preferable when the number of features is small.

\textit{Federated tabular synthesis} arises when clients hold different records rather than different features. \textbf{FedTabDiff} \cite{sattarov2024fedtabdiff} adapts FinDiff to this horizontally partitioned setting by training local diffusion models at clients and aggregating model updates through federated learning. This reduces raw-data sharing, but federated optimization alone does not provide a formal record-level privacy guarantee. \textbf{DP-Fed-FinDiff} \cite{sattarov2024differentially} therefore keeps the same basic federated diffusion setup but adds differential privacy to parameter updates through clipping and Gaussian noise. Its main lesson is that data locality and individual-level privacy are different requirements: federated learning controls where data are stored, whereas differential privacy controls how much one record can influence the learned generator.

\textit{Centralized differentially private synthesis} raises a different difficulty. \textbf{DP-FinDiff} \cite{sattarov2025privacy} studies differentially private diffusion for mixed-type tables when data are centrally available but record-level influence must still be bounded. Directly applying differentially private stochastic gradient descent (DP-SGD) is difficult because one-hot categorical representations increase dimensionality and sparsity, while clipping and Gaussian noise can distort the denoising signal unevenly across timesteps and features. DP-FinDiff therefore follows FinDiff by using dense categorical embeddings and training the denoiser in the embedding space under per-example clipping, Gaussian noise, and privacy random variable (PRV) accounting. It further uses adaptive timestep sampling and a feature-aggregated loss to reduce clipping-induced bias. This gives a practical differentially private variant of FinDiff, but it still faces the usual privacy--utility tradeoff, and formal differential privacy does not by itself address minority coverage, fairness, or bias inherited from the training data.

\textit{Fairness-preserving synthesis} broadens trustworthiness beyond privacy. \textbf{FairTabDDPM} \cite{yang2024balanced} starts from the observation that tabular diffusion models may reproduce group bias in the training data. To mitigate this, it builds balanced joint distributions of sensitive attributes and target labels, and conditions a TabDDPM-style Gaussian/multinomial diffusion model on both label and sensitive columns. This makes fairness control part of the generation process rather than a purely post-hoc filter. However, the guarantee is tied to the chosen sensitive attributes, target labels, and fairness metrics; reducing measured disparities does not necessarily remove other biased correlations or ensure fairness for every downstream use.

\textit{Memorization auditing} is another part of trustworthy synthesis, because a high-fidelity generator may reproduce individual training records too closely even without an explicit privacy failure. Fang et al. \cite{fang2025understanding} show that tabular diffusion models can memorize training samples and propose a mixed-type relative distance ratio to identify generated records that are much closer to their nearest training neighbor than to other training records. They further introduce \textbf{TabCutMix}, which swaps feature segments between same-class samples, and \textbf{TabCutMixPlus}, which swaps correlated feature groups to better preserve feature coherence. This line of work is useful because common distance-to-closest-record diagnostics can miss sample-level overfitting, but feature swapping may also create semantically invalid combinations in sensitive domains.

A follow-up study shifts the focus from model-level memorization to data-centric auditing \cite{fang2025closer}. It finds that memorized generations are often driven by a small subset of high-risk training records, and proposes \textbf{DynamicCut} to identify such records during a warm-up phase before retraining; \textbf{DynamicCutMix} combines this filtering with TabCutMix. The broader lesson is that privacy in tabular diffusion should be studied at three levels: formal mechanisms such as differential privacy, model-level attacks or memorization metrics, and data-centric auditing of repeatedly reproduced records. The remaining tradeoff is that removing high-risk records can also remove rare but analytically important cases, so memorization mitigation should be evaluated together with minority coverage, fairness, and domain validity.

\textbf{Comparative Analysis}:
The methods in Table~\ref{SummaryTrustDataSythesis} are best distinguished by their trust assumptions before considering the diffusion backbone. Cross-silo and federated synthesis reduce direct data sharing, but neither setting by itself gives a record-level privacy guarantee. Differentially private variants make individual influence explicit through a privacy budget, at the cost of utility, fidelity, or minority coverage. Fairness-aware synthesis changes the conditioning distribution to reduce measured group disparities, but it does not remove all biased correlations. Memorization auditing targets another failure mode: a non-private generator may reproduce particular training records even when aggregate fidelity and downstream utility look strong.  These differences imply that trustworthy tabular generation cannot be evaluated by a single fidelity or utility score. Studies should report the deployment setting, adversary or privacy definition, privacy budget or attack protocol, fairness metrics, minority-group utility, and sample-level memorization diagnostics together. Flow matching remains largely unexplored in this setting. Its faster sampling may affect utility, privacy leakage, or memorization behavior, but without explicit threat models and attack evaluations, efficiency should not be taken as evidence of trustworthiness.

\section{Diffusion and Flow Matching Models for Anomaly Detection}
\label{sec:AnoDec}

Anomaly detection aims to train diffusion or flow-matching models to learn the ``normal'' distribution of data from the training set and identify anomalies as deviations from this learned distribution or learned transport behavior in the test data. Formally, it is defined below (see Problem~\ref{prob:TabularAnomalyDetection}). Because these models can either score deviations directly or synthesize auxiliary samples for downstream fraud classifiers, we summarize representative methods in Table~\ref{SummaryAnomalyDetection} and organize the discussion into diffusion-based scoring (see Sec.~\ref{subsubsec:DiffusionScoringAnomalyDetection}), flow-matching-based scoring (see Sec.~\ref{subsubsec:FlowMatchingScoringAnomalyDetection}), auxiliary data generation for fraud detection (see Sec.~\ref{subsubsec:AuxiliaryFraudDetection}), and a final comparative analysis (see Sec.~\ref{subsubsec:AnomalyComparativeAnalysis}).

\begin{problem} [Tabular Anomaly Detection]\label{prob:TabularAnomalyDetection}
Given a $D$-dimensional training dataset $\mathbf{X}_{\text{train}} = \{\prescript{}{i}{\mathbf{x}}\}_{i=1}^{N_{\text{train}}}$, where each data point $\prescript{}{i}{\mathbf{x}} \in \mathbb{R}^{D}$ represents a record with $D$ features, the objective is to learn a function $f(\cdot): \mathbb{R}^{D} \to \{0, 1\}$ that distinguishes between normal and anomalous data points. During testing, given a $D$-dimensional test dataset $\mathbf{X}_{\text{test}} = \{\prescript{}{j}{\mathbf{x}}\}_{j=1}^{N_{\text{test}}}$, the function $f(\cdot)$ predicts $f(\prescript{}{j}{\mathbf{x}}) = 1$ if $\prescript{}{j}{\mathbf{x}}$ is an anomaly and $f(\prescript{}{j}{\mathbf{x}}) = 0$ if it is normal. Anomalies are identified as data points in $\mathbf{X}_{\text{test}}$ that significantly deviate from the normal patterns learned from $\mathbf{X}_{\text{train}}$.
\end{problem}

\begin{table}[h]
\centering
\caption{Overview of Diffusion and Flow Matching Models for Anomaly Detection}
\label{SummaryAnomalyDetection}
\resizebox{\columnwidth}{!}{%
\begin{tabular}{@{}llllll@{}}
\toprule
\textbf{Paradigm} & \textbf{Model Name} & \textbf{Year} & \textbf{Venue} & \textbf{Feature Type} & \textbf{Domain}\\ 
\midrule
\multirow{7}{*}{Diffusion} & \textbf{TabADM} \cite{zamberg2023tabadm}  & 2023         & ArXiv & Num by default & Generic\\
& \textbf{DTE} \cite{livernoche2024diffusion}  & 2024         & ICLR & Num by default & Generic\\
& \textbf{SDAD} \cite{li2024self}  & 2024         & Inf. Sci. & Num by default & Generic\\
& \textbf{NSCBAD} \cite{anonymous2024anomaly}  & 2024         & OpenReview & Num by default & Generic\\
& \textbf{FraudDiffuse} 
\cite{roy2024frauddiffuse}  & 2024         & ICAIF & Num+Cat & Finance\\
& \textbf{FraudDDPM} \cite{pushkarenko2024synthetic}  & 2024         & ISIJ & Num+Cat & Finance\\
& \textbf{DDAE} \cite{sattarov2025diffusion}  & 2025         & KDD & Num by default & Generic\\
\hline
Flow Matching & \textbf{TCCM} \cite{li2025scalable}  & 2025         & NeurIPS & Num by default & Generic\\
\bottomrule
\end{tabular}%
}
\end{table}

\subsubsection{Diffusion-Based Scoring}
\label{subsubsec:ScoringAnomalyDetection}
\label{subsubsec:DiffusionScoringAnomalyDetection}

Early diffusion-based anomaly detectors use the denoising process as a scoring device rather than as a data-release mechanism. \textbf{TabADM} \cite{zamberg2023tabadm} is designed for unsupervised tabular anomaly detection when the training set may already contain anomalies. It trains a robust DDPM with a sample-rejection step that downweights high-loss samples, and then scores a test record by its denoising or reconstruction loss across timesteps. Samples far from the learned high-density region tend to produce larger losses. This makes TabADM a direct diffusion-based scoring method, but it inherits the cost of repeated denoising and provides limited explainability or missing-value support.

\textbf{DTE} \cite{livernoche2024diffusion} starts from the same concern about computational cost. Instead of learning a full reverse denoising model, it estimates the diffusion time associated with a noisy input sample. The intuition is that anomalous samples are farther from the normal data manifold and therefore correspond to larger estimated diffusion times. DTE can use non-parametric or neural estimators and also supports reconstruction-based interpretation through deterministic ODE flows. It is more efficient than full DDPM-style scoring, but it is mainly designed for numerical features and still depends on how well diffusion time separates normal and abnormal regions.

After TabADM and DTE, later work changes what is scored. \textbf{SDAD} \cite{li2024self} argues that generative anomaly detectors should not only reconstruct data, but also learn discriminative representations. It first uses self-supervised pretext tasks to train an encoder that separates normal and anomalous samples in latent space, and then fits a denoising diffusion model to the normal latent distribution. At inference, samples with large latent reconstruction errors are treated as anomalies. This makes SDAD more representation-aware than direct input-space reconstruction, but it is designed for the semi-supervised setting with normal-only training data, and its pseudo-anomaly construction still relies on Gaussian noise rather than real abnormal patterns. \textbf{NSCBAD} \cite{anonymous2024anomaly} further shifts the score from reconstruction to the learned score function. Instead of estimating the full data density or running a full reverse denoising chain, it trains a noising conditional score network and scores a test sample by how well the predicted score matches injected noise at a fixed timestep. This avoids the long sequential inference of DDPM-based detectors and is therefore faster, but it still assumes that score mismatch under numerical noising is a reliable proxy for abnormality.

\textbf{DDAE} \cite{sattarov2025diffusion} takes a more lightweight reconstruction route. Instead of training a full diffusion generator, it imports diffusion-style scheduled noise into a denoising autoencoder: inputs are corrupted at different timesteps, a timestep-conditioned encoder--decoder reconstructs the clean sample, and the cumulative reconstruction error becomes the anomaly score. Its contrastive variant further shapes the latent space by separating clean and noised samples. This design keeps the useful idea of timestep-dependent corruption while avoiding iterative reverse sampling, but its behavior depends on the noise schedule and it mainly treats tabular inputs as standardized numerical inputs.

\subsubsection{Flow-Matching-Based Scoring}
\label{subsubsec:FlowMatchingScoringAnomalyDetection}

\textbf{TCCM} \cite{li2025scalable} pushes the simplification further from a flow-matching perspective. It trains only on normal samples and learns a time-conditioned contraction field toward the origin, using a fixed contraction target rather than a full probability-flow trajectory. At inference, it measures how far the predicted vector deviates from the expected contraction, so the anomaly score is computed in one forward pass instead of through multi-step denoising or ODE sampling. Because the residual remains in the input feature space, it also gives feature-wise attributions. The main caveat is the semi-supervised mismatch assumption: anomalous samples should fail to follow the contraction dynamics learned from normal data. Thus, its behavior still needs validation for mixed-type tables and high-stakes domains beyond standardized numerical benchmarks.

\begin{table*}[h]
\centering
\caption{Overview of Diffusion and Flow Matching Models for Tabular Data. From top to bottom, row shading marks \taskshade{OverallAugGenericBg}{data augmentation}, \taskshade{OverallImpBg}{data imputation}, \taskshade{OverallTrustBg}{trustworthy data synthesis}, and \taskshade{OverallAnomBg}{anomaly detection}. ``Num'' and ``Cat'' report numerical/categorical support and the main normalization or encoding: MM=min--max, QT=quantile transform, ZS=$Z$-score, OH=one-hot, AB=analog bits, DC=dictionary, DE=dummy, IE=integer, OE=ordinal, PSK=PSK, FT=feature tokenization or learned embedding, CC=CatConverter, Emb=learned embedding. ``\#Datasets'' is the number of benchmark datasets; ``?/57 A'' denotes the subset used from ADBench~\cite{han2022ADBench}. GitHub icons link to code; ``Complete'' indicates complete-data training.}
\label{tab:OverallSummary}
\resizebox{\linewidth}{!}{
\begin{tabular}{@{}llclllllrlccl@{}}
\toprule
\textbf{Paradigm} & \textbf{Model Name} & \textbf{Year} & \textbf{Venue} & \textbf{Tasks} & \textbf{Num} & \textbf{Cat} & \textbf{Backbone} & \#\textbf{Datasets} & \textbf{Metrics} & \textbf{Code} & \textbf{Complete} & \textbf{Domain}\\
\midrule
\multirow{14}{*}{Diffusion} & \cellcolor{OverallAugGenericBg}\textbf{SOS} \cite{kim2022sos} & \cellcolor{OverallAugGenericBg}2022 & \cellcolor{OverallAugGenericBg}KDD & \cellcolor{OverallAugGenericBg}Synthesis (single, generic) & \cellcolor{OverallAugGenericBg}\checkmark & \cellcolor{OverallAugGenericBg}\ding{55} & \cellcolor{OverallAugGenericBg}SDEs & \cellcolor{OverallAugGenericBg}6 & \cellcolor{OverallAugGenericBg}Utility & \cellcolor{OverallAugGenericBg}\codeurl{https://github.com/jayoungkim408/SOS} & \cellcolor{OverallAugGenericBg}\checkmark & \cellcolor{OverallAugGenericBg}Generic\\
 & \cellcolor{OverallAugGenericBg}\textbf{STaSy} \cite{kimstasy} & \cellcolor{OverallAugGenericBg}2023 & \cellcolor{OverallAugGenericBg}ICLR & \cellcolor{OverallAugGenericBg}Synthesis (single, generic) & \cellcolor{OverallAugGenericBg}\checkmark (MM) & \cellcolor{OverallAugGenericBg}\checkmark (OH) & \cellcolor{OverallAugGenericBg}SDEs & \cellcolor{OverallAugGenericBg}15 & \cellcolor{OverallAugGenericBg}Fidelity, Utility, Diversity & \cellcolor{OverallAugGenericBg}\codeurl{https://github.com/JayoungKim408/STaSy} & \cellcolor{OverallAugGenericBg}\checkmark & \cellcolor{OverallAugGenericBg}Generic\\
 & \cellcolor{OverallAugGenericBg}\textbf{TabDDPM} \cite{kotelnikov2023tabddpm} & \cellcolor{OverallAugGenericBg}2023 & \cellcolor{OverallAugGenericBg}ICML & \cellcolor{OverallAugGenericBg}Synthesis (single, generic) & \cellcolor{OverallAugGenericBg}\checkmark (QT) & \cellcolor{OverallAugGenericBg}\checkmark (OH) & \cellcolor{OverallAugGenericBg}DDPM+MLD & \cellcolor{OverallAugGenericBg}16 & \cellcolor{OverallAugGenericBg}Fidelity, Utility, Privacy & \cellcolor{OverallAugGenericBg}\codeurl{https://github.com/yandex-research/tab-ddpm} & \cellcolor{OverallAugGenericBg}\checkmark & \cellcolor{OverallAugGenericBg}Generic\\
 & \cellcolor{OverallAugGenericBg}\textbf{CoDi} \cite{lee2023codi} & \cellcolor{OverallAugGenericBg}2023 & \cellcolor{OverallAugGenericBg}ICML & \cellcolor{OverallAugGenericBg}Synthesis (single, generic) & \cellcolor{OverallAugGenericBg}\checkmark (MM) & \cellcolor{OverallAugGenericBg}\checkmark (OH) & \cellcolor{OverallAugGenericBg}DDPM+MLD & \cellcolor{OverallAugGenericBg}15 & \cellcolor{OverallAugGenericBg}Utility, Diversity & \cellcolor{OverallAugGenericBg}\codeurl{https://github.com/chaejeonglee/codi} & \cellcolor{OverallAugGenericBg}\checkmark & \cellcolor{OverallAugGenericBg}Generic\\ 
 & \cellcolor{OverallAugGenericBg}\textbf{AutoDiff} \cite{suh2023autodiff} & \cellcolor{OverallAugGenericBg}2023 & \cellcolor{OverallAugGenericBg}NeurIPSW & \cellcolor{OverallAugGenericBg}Synthesis (single, generic) & \cellcolor{OverallAugGenericBg}\checkmark & \cellcolor{OverallAugGenericBg}\checkmark & \cellcolor{OverallAugGenericBg}Any & \cellcolor{OverallAugGenericBg}15 & \cellcolor{OverallAugGenericBg}Fidelity, Utility, Privacy & \cellcolor{OverallAugGenericBg}\codeurl{https://github.com/ucla-trustworthy-ai-lab/autodiffusion} & \cellcolor{OverallAugGenericBg}\checkmark & \cellcolor{OverallAugGenericBg}Generic\\ 
 & \cellcolor{OverallAugGenericBg}\textbf{MissDiff} \cite{ouyang2023missdiff} & \cellcolor{OverallAugGenericBg}2023 & \cellcolor{OverallAugGenericBg}ICMLW & \cellcolor{OverallAugGenericBg}Synthesis (single, generic) & \cellcolor{OverallAugGenericBg}\checkmark (MM) & \cellcolor{OverallAugGenericBg}\checkmark (OH) & \cellcolor{OverallAugGenericBg}SDEs & \cellcolor{OverallAugGenericBg}3 & \cellcolor{OverallAugGenericBg}Fidelity, Utility & \cellcolor{OverallAugGenericBg}\ding{55} & \cellcolor{OverallAugGenericBg}\ding{55} & \cellcolor{OverallAugGenericBg}Generic\\
 & \cellcolor{OverallAugGenericBg}\textbf{TabSyn} \cite{zhangmixed} & \cellcolor{OverallAugGenericBg}2024 & \cellcolor{OverallAugGenericBg}ICLR & \cellcolor{OverallAugGenericBg}Synthesis (single, generic), Imputation & \cellcolor{OverallAugGenericBg}\checkmark & \cellcolor{OverallAugGenericBg}\checkmark (OH) & \cellcolor{OverallAugGenericBg}SDEs & \cellcolor{OverallAugGenericBg}6 & \cellcolor{OverallAugGenericBg}Fidelity, Utility, Diversity, Privacy & \cellcolor{OverallAugGenericBg}\codeurl{https://github.com/amazon-science/tabsyn} & \cellcolor{OverallAugGenericBg}\checkmark & \cellcolor{OverallAugGenericBg}Generic\\
 & \cellcolor{OverallAugGenericBg}\textbf{Forest-Diffusion} \cite{jolicoeur2024generating} & \cellcolor{OverallAugGenericBg}2024 & \cellcolor{OverallAugGenericBg}AISTATS & \cellcolor{OverallAugGenericBg}Synthesis (single, generic), Imputation & \cellcolor{OverallAugGenericBg}\checkmark & \cellcolor{OverallAugGenericBg}\checkmark (DE) & \cellcolor{OverallAugGenericBg}SDEs & \cellcolor{OverallAugGenericBg}27 & \cellcolor{OverallAugGenericBg}Fidelity, Diversity, Utility & \cellcolor{OverallAugGenericBg}\codeurl{https://github.com/SamsungSAILMontreal/ForestDiffusion} & \cellcolor{OverallAugGenericBg}\ding{55} & \cellcolor{OverallAugGenericBg}Generic\\
 & \cellcolor{OverallAugGenericBg}\textbf{TabDiff} \cite{shi2024tabdiff} & \cellcolor{OverallAugGenericBg}2025 & \cellcolor{OverallAugGenericBg}ICLR & \cellcolor{OverallAugGenericBg}Synthesis (single, generic), Imputation & \cellcolor{OverallAugGenericBg}\checkmark (MM) & \cellcolor{OverallAugGenericBg}\checkmark (OH) & \cellcolor{OverallAugGenericBg}SDEs+MSD & \cellcolor{OverallAugGenericBg}7 & \cellcolor{OverallAugGenericBg}Fidelity, Diversity, Utility & \cellcolor{OverallAugGenericBg}\codeurl{https://github.com/MinkaiXu/TabDiff} & \cellcolor{OverallAugGenericBg}\checkmark & \cellcolor{OverallAugGenericBg}Generic\\
 & \cellcolor{OverallAugGenericBg}\textbf{TabUnite} \cite{si2024tabunite} & \cellcolor{OverallAugGenericBg}2024 & \cellcolor{OverallAugGenericBg}OpenReview & \cellcolor{OverallAugGenericBg}Synthesis (single, generic) & \cellcolor{OverallAugGenericBg}\checkmark (QT) & \cellcolor{OverallAugGenericBg}\checkmark (AB, PSK, DC) & \cellcolor{OverallAugGenericBg}SDEs+MSD & \cellcolor{OverallAugGenericBg}10 & \cellcolor{OverallAugGenericBg}Fidelity, Diversity, Utility & \cellcolor{OverallAugGenericBg}\codeurl{https://github.com/jacobyhsi/TabUnite} & \cellcolor{OverallAugGenericBg}\checkmark & \cellcolor{OverallAugGenericBg}Generic\\
 & \cellcolor{OverallAugGenericBg}\textbf{CDTD} \cite{mueller2024continuous} & \cellcolor{OverallAugGenericBg}2025 & \cellcolor{OverallAugGenericBg}ICLR & \cellcolor{OverallAugGenericBg}Synthesis (single, generic) & \cellcolor{OverallAugGenericBg}\checkmark & \cellcolor{OverallAugGenericBg}\checkmark (FT) & \cellcolor{OverallAugGenericBg}Latent SDEs & \cellcolor{OverallAugGenericBg}11 & \cellcolor{OverallAugGenericBg}Fidelity, Utility, Privacy & \cellcolor{OverallAugGenericBg}\codeurl{https://github.com/muellermarkus/cdtd} & \cellcolor{OverallAugGenericBg}\checkmark & \cellcolor{OverallAugGenericBg}Generic\\
 & \cellcolor{OverallAugGenericBg}\textbf{TabRep} \cite{si2026tabrep} & \cellcolor{OverallAugGenericBg}2026 & \cellcolor{OverallAugGenericBg}TMLR & \cellcolor{OverallAugGenericBg}Synthesis (single, generic) & \cellcolor{OverallAugGenericBg}\checkmark (QT) & \cellcolor{OverallAugGenericBg}\checkmark (CC) & \cellcolor{OverallAugGenericBg}DDPM/FM & \cellcolor{OverallAugGenericBg}7 & \cellcolor{OverallAugGenericBg}Fidelity, Utility, Privacy & \cellcolor{OverallAugGenericBg}\codeurl{https://github.com/jacobyhsi/TabRep} & \cellcolor{OverallAugGenericBg}\checkmark & \cellcolor{OverallAugGenericBg}Generic\\
 & \cellcolor{OverallAugGenericBg}\textbf{TabNAT} \cite{zhang2025tabnat} & \cellcolor{OverallAugGenericBg}2025 & \cellcolor{OverallAugGenericBg}ICML & \cellcolor{OverallAugGenericBg}Synthesis (single, generic), Imputation & \cellcolor{OverallAugGenericBg}\checkmark & \cellcolor{OverallAugGenericBg}\checkmark (Emb) & \cellcolor{OverallAugGenericBg}Conditional DDPM & \cellcolor{OverallAugGenericBg}10 & \cellcolor{OverallAugGenericBg}Fidelity, Utility, Privacy, Runtime & \cellcolor{OverallAugGenericBg}\codeurl{https://github.com/fangliancheng/TabNAT} & \cellcolor{OverallAugGenericBg}\checkmark & \cellcolor{OverallAugGenericBg}Generic\\
 & \cellcolor{OverallAugGenericBg}\textbf{TabuSDE} \cite{malashin2025stochastic} & \cellcolor{OverallAugGenericBg}2025 & \cellcolor{OverallAugGenericBg}ICDMW & \cellcolor{OverallAugGenericBg}Synthesis (single, generic), Imputation & \cellcolor{OverallAugGenericBg}\checkmark & \cellcolor{OverallAugGenericBg}\checkmark (Emb) & \cellcolor{OverallAugGenericBg}SDE & \cellcolor{OverallAugGenericBg}25 & \cellcolor{OverallAugGenericBg}Fidelity, Imputation, Privacy & \cellcolor{OverallAugGenericBg}\ding{55} & \cellcolor{OverallAugGenericBg}\ding{55} & \cellcolor{OverallAugGenericBg}Generic\\
\cmidrule(lr){1-13}
\multirow{4}{*}{Flow Matching} & \cellcolor{OverallAugGenericBg}\textbf{TabbyFlow} \cite{guzman-cordero2025exponential} & \cellcolor{OverallAugGenericBg}2025 & \cellcolor{OverallAugGenericBg}ICML & \cellcolor{OverallAugGenericBg}Synthesis (single, generic) & \cellcolor{OverallAugGenericBg}\checkmark & \cellcolor{OverallAugGenericBg}\checkmark & \cellcolor{OverallAugGenericBg}VFM & \cellcolor{OverallAugGenericBg}6 & \cellcolor{OverallAugGenericBg}Fidelity, Diversity, Utility, Privacy & \cellcolor{OverallAugGenericBg}\codeurl{https://github.com/andresguzco/ef-vfm} & \cellcolor{OverallAugGenericBg}\checkmark & \cellcolor{OverallAugGenericBg}Generic\\
 & \cellcolor{OverallAugGenericBg}\textbf{TabSynFlow} \cite{nasution2026flow} & \cellcolor{OverallAugGenericBg}2026 & \cellcolor{OverallAugGenericBg}TMLR & \cellcolor{OverallAugGenericBg}Synthesis (single, generic) & \cellcolor{OverallAugGenericBg}\checkmark & \cellcolor{OverallAugGenericBg}\checkmark & \cellcolor{OverallAugGenericBg}Latent Flow Matching & \cellcolor{OverallAugGenericBg}6 & \cellcolor{OverallAugGenericBg}Utility, Privacy, Runtime & \cellcolor{OverallAugGenericBg}\codeurl{https://github.com/rulnasution/tabular-flow-matching} & \cellcolor{OverallAugGenericBg}\checkmark & \cellcolor{OverallAugGenericBg}Generic\\
 & \cellcolor{OverallAugGenericBg}\textbf{TabFlowM} \cite{anonymous2026tabflowm} & \cellcolor{OverallAugGenericBg}2026 & \cellcolor{OverallAugGenericBg}TMLR Sub. & \cellcolor{OverallAugGenericBg}Synthesis (single, generic) & \cellcolor{OverallAugGenericBg}\checkmark & \cellcolor{OverallAugGenericBg}\checkmark (FT) & \cellcolor{OverallAugGenericBg}Latent Flow Matching & \cellcolor{OverallAugGenericBg}12 & \cellcolor{OverallAugGenericBg}Fidelity, Utility, Privacy, Runtime & \cellcolor{OverallAugGenericBg}\ding{55} & \cellcolor{OverallAugGenericBg}\checkmark & \cellcolor{OverallAugGenericBg}Generic\\
 & \cellcolor{OverallAugGenericBg}\textbf{TabCascade} \cite{mueller2026cascaded} & \cellcolor{OverallAugGenericBg}2026 & \cellcolor{OverallAugGenericBg}ArXiv & \cellcolor{OverallAugGenericBg}Synthesis (single, generic) & \cellcolor{OverallAugGenericBg}\checkmark & \cellcolor{OverallAugGenericBg}\checkmark & \cellcolor{OverallAugGenericBg}Conditional Flow Matching & \cellcolor{OverallAugGenericBg}12 & \cellcolor{OverallAugGenericBg}Fidelity, Utility, Privacy & \cellcolor{OverallAugGenericBg}\codeurl{https://github.com/muellermarkus/tabcascade} & \cellcolor{OverallAugGenericBg}\ding{55} & \cellcolor{OverallAugGenericBg}Generic\\
\cmidrule(lr){1-13}
\multirow{6}{*}{Diffusion} & \cellcolor{OverallAugHealthBg}\textbf{MedDiff} \cite{he2023meddiff} & \cellcolor{OverallAugHealthBg}2023 & \cellcolor{OverallAugHealthBg}ArXiv & \cellcolor{OverallAugHealthBg}Synthesis (single, healthcare) & \cellcolor{OverallAugHealthBg}\checkmark & \cellcolor{OverallAugHealthBg}\ding{55} & \cellcolor{OverallAugHealthBg}DDIM & \cellcolor{OverallAugHealthBg}2 & \cellcolor{OverallAugHealthBg}Fidelity, Utility & \cellcolor{OverallAugHealthBg}\ding{55} & \cellcolor{OverallAugHealthBg}\checkmark & \cellcolor{OverallAugHealthBg}Healthcare\\
 & \cellcolor{OverallAugHealthBg}\textbf{EHR-TabDDPM} \cite{ceritli2023synthesizing} & \cellcolor{OverallAugHealthBg}2023 & \cellcolor{OverallAugHealthBg}ArXiv & \cellcolor{OverallAugHealthBg}Synthesis (single, healthcare) & \cellcolor{OverallAugHealthBg}\checkmark (QT) & \cellcolor{OverallAugHealthBg}\checkmark (OH) & \cellcolor{OverallAugHealthBg}DDPM & \cellcolor{OverallAugHealthBg}4 & \cellcolor{OverallAugHealthBg}Fidelity, Utility, Privacy & \cellcolor{OverallAugHealthBg}\ding{55} & \cellcolor{OverallAugHealthBg}\checkmark & \cellcolor{OverallAugHealthBg}Healthcare\\
 & \cellcolor{OverallAugHealthBg}\textbf{DPM-EHR}* \cite{nicholas2023synthetic} & \cellcolor{OverallAugHealthBg}2023 & \cellcolor{OverallAugHealthBg}NeurIPSW & \cellcolor{OverallAugHealthBg}Synthesis (single, healthcare) & \cellcolor{OverallAugHealthBg}\checkmark & \cellcolor{OverallAugHealthBg}\checkmark (OH) & \cellcolor{OverallAugHealthBg}DDPM & \cellcolor{OverallAugHealthBg}2 & \cellcolor{OverallAugHealthBg}Fidelity, Diversity, Utility, Privacy & \cellcolor{OverallAugHealthBg}\ding{55} & \cellcolor{OverallAugHealthBg}\checkmark & \cellcolor{OverallAugHealthBg}Healthcare\\
 & \cellcolor{OverallAugHealthBg}\textbf{FlexGen-EHR} \cite{he2024flexible} & \cellcolor{OverallAugHealthBg}2024 & \cellcolor{OverallAugHealthBg}ICLR & \cellcolor{OverallAugHealthBg}Synthesis (single, healthcare) & \cellcolor{OverallAugHealthBg}\checkmark + TS (FT) & \cellcolor{OverallAugHealthBg}\checkmark (FT) & \cellcolor{OverallAugHealthBg}DDPM & \cellcolor{OverallAugHealthBg}2 & \cellcolor{OverallAugHealthBg}Fidelity, Utility, Privacy & \cellcolor{OverallAugHealthBg}\ding{55} & \cellcolor{OverallAugHealthBg}\ding{55} & \cellcolor{OverallAugHealthBg}Healthcare\\
 & \cellcolor{OverallAugHealthBg}\textbf{EHRDiff} \cite{yuan2024ehrdiff} & \cellcolor{OverallAugHealthBg}2024 & \cellcolor{OverallAugHealthBg}TMLR & \cellcolor{OverallAugHealthBg}Synthesis (single, healthcare) & \cellcolor{OverallAugHealthBg}\checkmark (MM) & \cellcolor{OverallAugHealthBg}\checkmark (OH) & \cellcolor{OverallAugHealthBg}SDEs & \cellcolor{OverallAugHealthBg}3 & \cellcolor{OverallAugHealthBg}Fidelity, Utility, Privacy & \cellcolor{OverallAugHealthBg}\codeurl{https://github.com/sczzz3/ehrdiff} & \cellcolor{OverallAugHealthBg}\checkmark & \cellcolor{OverallAugHealthBg}Healthcare \\
 & \cellcolor{OverallAugHealthBg}\textbf{EHR-D3PM} \cite{han2024guided} & \cellcolor{OverallAugHealthBg}2025 & \cellcolor{OverallAugHealthBg}TMLR & \cellcolor{OverallAugHealthBg}Synthesis (single, healthcare) & \cellcolor{OverallAugHealthBg}\ding{55} & \cellcolor{OverallAugHealthBg}\checkmark (OH) & \cellcolor{OverallAugHealthBg}D3PM & \cellcolor{OverallAugHealthBg}3 & \cellcolor{OverallAugHealthBg}Fidelity, Utility, Privacy & \cellcolor{OverallAugHealthBg}\ding{55} & \cellcolor{OverallAugHealthBg}\checkmark & \cellcolor{OverallAugHealthBg}Healthcare \\
\cmidrule(lr){1-13}
Flow Matching & \cellcolor{OverallAugHealthBg}\textbf{PatientFlow} \cite{branco2026patientflow} & \cellcolor{OverallAugHealthBg}2026 & \cellcolor{OverallAugHealthBg}AI Med. & \cellcolor{OverallAugHealthBg}Synthesis (single, healthcare) & \cellcolor{OverallAugHealthBg}\checkmark + TS & \cellcolor{OverallAugHealthBg}\checkmark (Emb) & \cellcolor{OverallAugHealthBg}Latent Flow Matching & \cellcolor{OverallAugHealthBg}1 & \cellcolor{OverallAugHealthBg}Fidelity, Utility, Clinical Rules, Privacy & \cellcolor{OverallAugHealthBg}\codeurl{https://github.com/RubenBranco/PatientFlow} & \cellcolor{OverallAugHealthBg}\ding{55} & \cellcolor{OverallAugHealthBg}Healthcare\\
\cmidrule(lr){1-13}
\multirow{3}{*}{Diffusion} & \cellcolor{OverallAugFinanceBg}\textbf{FinDiff} \cite{sattarov2023findiff} & \cellcolor{OverallAugFinanceBg}2023 & \cellcolor{OverallAugFinanceBg}ICAIF & \cellcolor{OverallAugFinanceBg}Synthesis (single, finance) & \cellcolor{OverallAugFinanceBg}\checkmark (ZS) & \cellcolor{OverallAugFinanceBg}\checkmark (OH) & \cellcolor{OverallAugFinanceBg}DDPM & \cellcolor{OverallAugFinanceBg}3 & \cellcolor{OverallAugFinanceBg}Fidelity, Utility, Privacy & \cellcolor{OverallAugFinanceBg}\codeurl{https://github.com/sattarov/FinDiff} & \cellcolor{OverallAugFinanceBg}\checkmark & \cellcolor{OverallAugFinanceBg}Finance \\
 & \cellcolor{OverallAugFinanceBg}\textbf{EntTabDiff} \cite{liu2024entity} & \cellcolor{OverallAugFinanceBg}2024 & \cellcolor{OverallAugFinanceBg}ICAIF & \cellcolor{OverallAugFinanceBg}Synthesis (single, finance) & \cellcolor{OverallAugFinanceBg}\checkmark (ZS) & \cellcolor{OverallAugFinanceBg}\checkmark (OH) & \cellcolor{OverallAugFinanceBg}DDPM & \cellcolor{OverallAugFinanceBg}3 & \cellcolor{OverallAugFinanceBg}Fidelity, Utility, Privacy & \cellcolor{OverallAugFinanceBg}\ding{55} & \cellcolor{OverallAugFinanceBg}\checkmark & \cellcolor{OverallAugFinanceBg}Finance \\
 & \cellcolor{OverallAugFinanceBg}\textbf{Imb-FinDiff} \cite{schreyer2024imb} & \cellcolor{OverallAugFinanceBg}2024 & \cellcolor{OverallAugFinanceBg}ICAIF & \cellcolor{OverallAugFinanceBg}Synthesis (single, finance) & \cellcolor{OverallAugFinanceBg}\checkmark (ZS) & \cellcolor{OverallAugFinanceBg}\checkmark (OH) & \cellcolor{OverallAugFinanceBg}DDPM & \cellcolor{OverallAugFinanceBg}4 & \cellcolor{OverallAugFinanceBg}Fidelity, Utility & \cellcolor{OverallAugFinanceBg}\ding{55} & \cellcolor{OverallAugFinanceBg}\checkmark & \cellcolor{OverallAugFinanceBg}Finance \\
\cmidrule(lr){1-13}
\multirow{2}{*}{Diffusion} & \cellcolor{OverallAugRelBg}\textbf{ClavaDDPM} \cite{pang2024clavaddpm} & \cellcolor{OverallAugRelBg}2024 & \cellcolor{OverallAugRelBg}NeurIPS & \cellcolor{OverallAugRelBg}Synthesis (multi-relational) & \cellcolor{OverallAugRelBg}\checkmark & \cellcolor{OverallAugRelBg}\checkmark(IE) & \cellcolor{OverallAugRelBg}DDPM & \cellcolor{OverallAugRelBg}5 & \cellcolor{OverallAugRelBg}Fidelity, Diversity, Utility, Dependency & \cellcolor{OverallAugRelBg}\codeurl{https://github.com/weipang142857/clavaddpm} & \cellcolor{OverallAugRelBg}\checkmark & \cellcolor{OverallAugRelBg}Generic \\
 & \cellcolor{OverallAugRelBg}\textbf{GNN-TabSyn} \cite{hudovernik2024relational} & \cellcolor{OverallAugRelBg}2024 & \cellcolor{OverallAugRelBg}NeurIPSW & \cellcolor{OverallAugRelBg}Synthesis (multi-relational) & \cellcolor{OverallAugRelBg}\checkmark & \cellcolor{OverallAugRelBg}\checkmark(IE) & \cellcolor{OverallAugRelBg}DDPM & \cellcolor{OverallAugRelBg}6 & \cellcolor{OverallAugRelBg}Fidelity, Utility, Privacy & \cellcolor{OverallAugRelBg}\codeurl{https://github.com/ValterH/relational-graph-conditioned-diffusion} & \cellcolor{OverallAugRelBg}\checkmark & \cellcolor{OverallAugRelBg}Generic \\
\midrule
\multirow{10}{*}{Diffusion} & \cellcolor{OverallImpBg}\textbf{TabCSDI} \cite{zheng2022diffusion} & \cellcolor{OverallImpBg}2022 & \cellcolor{OverallImpBg}NeurIPSW & \cellcolor{OverallImpBg}Imputation & \cellcolor{OverallImpBg}\checkmark & \cellcolor{OverallImpBg}\checkmark (OH, AB, FT) & \cellcolor{OverallImpBg}Conditional DDPM & \cellcolor{OverallImpBg}7 & \cellcolor{OverallImpBg}Accuracy & \cellcolor{OverallImpBg}\codeurl{https://github.com/pfnet-research/CSDI_T} & \cellcolor{OverallImpBg}\ding{55} & \cellcolor{OverallImpBg}Generic\\
 & \cellcolor{OverallImpBg}\textbf{TabDiff} \cite{shi2024tabdiff} & \cellcolor{OverallImpBg}2025 & \cellcolor{OverallImpBg}ICLR & \cellcolor{OverallImpBg}Imputation & \cellcolor{OverallImpBg}\checkmark & \cellcolor{OverallImpBg}\checkmark & \cellcolor{OverallImpBg}Conditional DDPM & \cellcolor{OverallImpBg}7 & \cellcolor{OverallImpBg}Accuracy & \cellcolor{OverallImpBg}\codeurl{https://github.com/MinkaiXu/TabDiff} & \cellcolor{OverallImpBg}\ding{55} & \cellcolor{OverallImpBg}Generic\\
 & \cellcolor{OverallImpBg}\textbf{SimpDM} \cite{liu2024self} & \cellcolor{OverallImpBg}2024 & \cellcolor{OverallImpBg}CIKM & \cellcolor{OverallImpBg}Imputation & \cellcolor{OverallImpBg}\checkmark (QT) & \cellcolor{OverallImpBg}\checkmark (OH) & \cellcolor{OverallImpBg}DDPM+MLD & \cellcolor{OverallImpBg}17 & \cellcolor{OverallImpBg}Accuracy & \cellcolor{OverallImpBg}\codeurl{https://github.com/yixinliu233/simpdm} & \cellcolor{OverallImpBg}\ding{55} & \cellcolor{OverallImpBg}Generic \\
 & \cellcolor{OverallImpBg}\textbf{MTabGen} \cite{villaizan2024diffusion} & \cellcolor{OverallImpBg}2025 & \cellcolor{OverallImpBg}TKDD & \cellcolor{OverallImpBg}Imputation & \cellcolor{OverallImpBg}\checkmark (QT) & \cellcolor{OverallImpBg}\checkmark (OE) & \cellcolor{OverallImpBg}DDPM+MLD & \cellcolor{OverallImpBg}10 & \cellcolor{OverallImpBg}Utility & \cellcolor{OverallImpBg}\ding{55} & \cellcolor{OverallImpBg}\ding{55} & \cellcolor{OverallImpBg}Generic \\
 & \cellcolor{OverallImpBg}\textbf{DDPM-Perlin} \cite{wibisono2024natural} & \cellcolor{OverallImpBg}2024 & \cellcolor{OverallImpBg}KBS & \cellcolor{OverallImpBg}Imputation & \cellcolor{OverallImpBg}\checkmark & \cellcolor{OverallImpBg}\ding{55} & \cellcolor{OverallImpBg}DDPM & \cellcolor{OverallImpBg}10 & \cellcolor{OverallImpBg}Accuracy & \cellcolor{OverallImpBg}\ding{55} & \cellcolor{OverallImpBg}\ding{55} & \cellcolor{OverallImpBg}Generic \\
 & \cellcolor{OverallImpBg}\textbf{NewImp} \cite{chen2024rethinking} & \cellcolor{OverallImpBg}2024 & \cellcolor{OverallImpBg}NeurIPS & \cellcolor{OverallImpBg}Imputation & \cellcolor{OverallImpBg}\checkmark & \cellcolor{OverallImpBg}\ding{55} & \cellcolor{OverallImpBg}Wasserstein GF & \cellcolor{OverallImpBg}8 & \cellcolor{OverallImpBg}MAE, WASS & \cellcolor{OverallImpBg}\codeurl{https://github.com/JustusvLiebig/NewImp} & \cellcolor{OverallImpBg}\ding{55} & \cellcolor{OverallImpBg}Generic \\
 & \cellcolor{OverallImpBg}\textbf{DiffPuter} \cite{zhang2024unleashing} & \cellcolor{OverallImpBg}2025 & \cellcolor{OverallImpBg}ICLR & \cellcolor{OverallImpBg}Imputation & \cellcolor{OverallImpBg}\checkmark & \cellcolor{OverallImpBg}\checkmark (OH) & \cellcolor{OverallImpBg}DDPM & \cellcolor{OverallImpBg}10 & \cellcolor{OverallImpBg}Accuracy & \cellcolor{OverallImpBg}\codeurl{https://github.com/hengruizhang98/DiffPuter} & \cellcolor{OverallImpBg}\ding{55} & \cellcolor{OverallImpBg}Generic \\
 & \cellcolor{OverallImpBg}\textbf{HARPOON} \cite{shankar2026harpoon} & \cellcolor{OverallImpBg}2026 & \cellcolor{OverallImpBg}ICLR & \cellcolor{OverallImpBg}Imputation, Conditional Generation & \cellcolor{OverallImpBg}\checkmark (ZS) & \cellcolor{OverallImpBg}\checkmark (OH) & \cellcolor{OverallImpBg}Conditional DDPM & \cellcolor{OverallImpBg}8 & \cellcolor{OverallImpBg}MSE, Accuracy, Violation, Fidelity, Utility, Runtime & \cellcolor{OverallImpBg}\codeurl{https://github.com/aditya-shankar-iiit/Harpoon} & \cellcolor{OverallImpBg}\checkmark & \cellcolor{OverallImpBg}Generic \\
 & \cellcolor{OverallImpBg}\textbf{TabNAT} \cite{zhang2025tabnat} & \cellcolor{OverallImpBg}2025 & \cellcolor{OverallImpBg}ICML & \cellcolor{OverallImpBg}Imputation & \cellcolor{OverallImpBg}\checkmark & \cellcolor{OverallImpBg}\checkmark (Emb) & \cellcolor{OverallImpBg}Conditional DDPM & \cellcolor{OverallImpBg}5 & \cellcolor{OverallImpBg}MAE, Accuracy & \cellcolor{OverallImpBg}\codeurl{https://github.com/fangliancheng/TabNAT} & \cellcolor{OverallImpBg}\checkmark & \cellcolor{OverallImpBg}Generic \\
 & \cellcolor{OverallImpBg}\textbf{TabuSDE} \cite{malashin2025stochastic} & \cellcolor{OverallImpBg}2025 & \cellcolor{OverallImpBg}ICDMW & \cellcolor{OverallImpBg}Imputation & \cellcolor{OverallImpBg}\checkmark & \cellcolor{OverallImpBg}\checkmark (Emb) & \cellcolor{OverallImpBg}SDE & \cellcolor{OverallImpBg}25 & \cellcolor{OverallImpBg}MSE, Accuracy, CPS & \cellcolor{OverallImpBg}\ding{55} & \cellcolor{OverallImpBg}\ding{55} & \cellcolor{OverallImpBg}Generic \\
\midrule
\multirow{7}{*}{Diffusion} & \cellcolor{OverallTrustBg}\textbf{SiloFuse} \cite{shankar2024silofuse} & \cellcolor{OverallTrustBg}2024 & \cellcolor{OverallTrustBg}ICDE & \cellcolor{OverallTrustBg}Trustworthy Synthesis & \cellcolor{OverallTrustBg}\checkmark (FT) & \cellcolor{OverallTrustBg}\checkmark (FT) & \cellcolor{OverallTrustBg}Latent DDPM & \cellcolor{OverallTrustBg}9 & \cellcolor{OverallTrustBg}Fidelity, Utility, Privacy & \cellcolor{OverallTrustBg}\ding{55} & \cellcolor{OverallTrustBg}\checkmark & \cellcolor{OverallTrustBg}Generic \\
 & \cellcolor{OverallTrustBg}\textbf{FedTabDiff} \cite{sattarov2024fedtabdiff} & \cellcolor{OverallTrustBg}2024 & \cellcolor{OverallTrustBg}ArXiv & \cellcolor{OverallTrustBg}Trustworthy Synthesis & \cellcolor{OverallTrustBg}\checkmark (QT) & \cellcolor{OverallTrustBg}\checkmark (FT) & \cellcolor{OverallTrustBg}DDPM & \cellcolor{OverallTrustBg}2 & \cellcolor{OverallTrustBg}Fidelity, Utility, Privacy & \cellcolor{OverallTrustBg}\codeurl{https://github.com/sattarov/fedtabdiff} & \cellcolor{OverallTrustBg}\checkmark & \cellcolor{OverallTrustBg}Generic \\
 & \cellcolor{OverallTrustBg}\textbf{FairTabDDPM}* \cite{yang2024balanced} & \cellcolor{OverallTrustBg}2025 & \cellcolor{OverallTrustBg}TMLR & \cellcolor{OverallTrustBg}Trustworthy Synthesis & \cellcolor{OverallTrustBg}\checkmark (QT) & \cellcolor{OverallTrustBg}\checkmark (OH) & \cellcolor{OverallTrustBg}DDPM & \cellcolor{OverallTrustBg}3 & \cellcolor{OverallTrustBg}Fidelity, Diversity, Utility, Privacy, Fairness & \cellcolor{OverallTrustBg}\codeurl{https://github.com/comp-well-org/fair-tab-diffusion} & \cellcolor{OverallTrustBg}\checkmark & \cellcolor{OverallTrustBg}Generic \\
 & \cellcolor{OverallTrustBg}\textbf{DP-Fed-FinDiff} \cite{sattarov2024differentially} & \cellcolor{OverallTrustBg}2024 & \cellcolor{OverallTrustBg}ArXiv & \cellcolor{OverallTrustBg}Trustworthy Synthesis & \cellcolor{OverallTrustBg}\checkmark (QT) & \cellcolor{OverallTrustBg}\checkmark (FT) & \cellcolor{OverallTrustBg}DDPM & \cellcolor{OverallTrustBg}4 & \cellcolor{OverallTrustBg}Fidelity, Utility, Privacy & \cellcolor{OverallTrustBg}\ding{55} & \cellcolor{OverallTrustBg}\checkmark & \cellcolor{OverallTrustBg}Finance \\
 & \cellcolor{OverallTrustBg}\textbf{DP-FinDiff} \cite{sattarov2025privacy} & \cellcolor{OverallTrustBg}2025 & \cellcolor{OverallTrustBg}ArXiv & \cellcolor{OverallTrustBg}Trustworthy Synthesis & \cellcolor{OverallTrustBg}\checkmark & \cellcolor{OverallTrustBg}\checkmark (Emb) & \cellcolor{OverallTrustBg}DP-DDPM & \cellcolor{OverallTrustBg}5 & \cellcolor{OverallTrustBg}Fidelity, Utility, Privacy & \cellcolor{OverallTrustBg}\codeurl{https://github.com/sattarov/FinDiff} & \cellcolor{OverallTrustBg}\checkmark & \cellcolor{OverallTrustBg}Generic \\
 & \cellcolor{OverallTrustBg}\textbf{TabCutMix/TabCutMixPlus} \cite{fang2025understanding} & \cellcolor{OverallTrustBg}2025 & \cellcolor{OverallTrustBg}ICML & \cellcolor{OverallTrustBg}Trustworthy Synthesis & \cellcolor{OverallTrustBg}\checkmark & \cellcolor{OverallTrustBg}\checkmark & \cellcolor{OverallTrustBg}-- & \cellcolor{OverallTrustBg}7 & \cellcolor{OverallTrustBg}Memorization, Fidelity, Utility & \cellcolor{OverallTrustBg}\codeurl{https://github.com/fangzy96/TabCutMix} & \cellcolor{OverallTrustBg}\checkmark & \cellcolor{OverallTrustBg}Generic \\
 & \cellcolor{OverallTrustBg}\textbf{DynamicCut} \cite{fang2025closer} & \cellcolor{OverallTrustBg}2026 & \cellcolor{OverallTrustBg}TMLR & \cellcolor{OverallTrustBg}Trustworthy Synthesis & \cellcolor{OverallTrustBg}\checkmark & \cellcolor{OverallTrustBg}\checkmark & \cellcolor{OverallTrustBg}-- & \cellcolor{OverallTrustBg}5 & \cellcolor{OverallTrustBg}Memorization, Fidelity, Utility & \cellcolor{OverallTrustBg}\codeurl{https://github.com/fangzy96/DynamicCut} & \cellcolor{OverallTrustBg}\checkmark & \cellcolor{OverallTrustBg}Generic \\
\midrule
\multirow{7}{*}{Diffusion} & \cellcolor{OverallAnomBg}\textbf{TabADM} \cite{zamberg2023tabadm} & \cellcolor{OverallAnomBg}2023 & \cellcolor{OverallAnomBg}ArXiv & \cellcolor{OverallAnomBg}Anomaly Detection & \cellcolor{OverallAnomBg}\checkmark & \cellcolor{OverallAnomBg}\ding{55} & \cellcolor{OverallAnomBg}DDPM & \cellcolor{OverallAnomBg}32/57 A & \cellcolor{OverallAnomBg}Accuracy & \cellcolor{OverallAnomBg}\ding{55} & \cellcolor{OverallAnomBg}\checkmark & \cellcolor{OverallAnomBg}Generic\\
 & \cellcolor{OverallAnomBg}\textbf{DTE} \cite{livernoche2024diffusion} & \cellcolor{OverallAnomBg}2024 & \cellcolor{OverallAnomBg}ICLR & \cellcolor{OverallAnomBg}Anomaly Detection & \cellcolor{OverallAnomBg}\checkmark (ZS) & \cellcolor{OverallAnomBg}\ding{55} & \cellcolor{OverallAnomBg}DDPM & \cellcolor{OverallAnomBg}57/57 A & \cellcolor{OverallAnomBg}Accuracy & \cellcolor{OverallAnomBg}\codeurl{https://github.com/vicliv/DTE} & \cellcolor{OverallAnomBg}\checkmark & \cellcolor{OverallAnomBg}Generic\\
 & \cellcolor{OverallAnomBg}\textbf{SDAD} \cite{li2024self} & \cellcolor{OverallAnomBg}2024 & \cellcolor{OverallAnomBg}Inf. Sci. & \cellcolor{OverallAnomBg}Anomaly Detection & \cellcolor{OverallAnomBg}\checkmark & \cellcolor{OverallAnomBg}\ding{55} & \cellcolor{OverallAnomBg}Latent DDPM & \cellcolor{OverallAnomBg}10/57 A & \cellcolor{OverallAnomBg}Accuracy & \cellcolor{OverallAnomBg}\ding{55} & \cellcolor{OverallAnomBg}\checkmark & \cellcolor{OverallAnomBg}Generic\\
 & \cellcolor{OverallAnomBg}\textbf{NSCBAD} \cite{anonymous2024anomaly} & \cellcolor{OverallAnomBg}2024 & \cellcolor{OverallAnomBg}OpenReview & \cellcolor{OverallAnomBg}Anomaly Detection & \cellcolor{OverallAnomBg}\checkmark & \cellcolor{OverallAnomBg}\ding{55} & \cellcolor{OverallAnomBg}SDEs & \cellcolor{OverallAnomBg}57/57 A+15 & \cellcolor{OverallAnomBg}Accuracy & \cellcolor{OverallAnomBg}\codeurl{https://openreview.net/forum?id=7QDIFrtAsB} & \cellcolor{OverallAnomBg}\checkmark & \cellcolor{OverallAnomBg}Generic \\
 & \cellcolor{OverallAnomBg}\textbf{FraudDiffuse} \cite{roy2024frauddiffuse} & \cellcolor{OverallAnomBg}2024 & \cellcolor{OverallAnomBg}ICAIF & \cellcolor{OverallAnomBg}Anomaly Detection, Synthesis (single, finance) & \cellcolor{OverallAnomBg}\checkmark & \cellcolor{OverallAnomBg}\checkmark (FT) & \cellcolor{OverallAnomBg}Latent DDPM & \cellcolor{OverallAnomBg}2 & \cellcolor{OverallAnomBg}Utility & \cellcolor{OverallAnomBg}\ding{55} & \cellcolor{OverallAnomBg}\checkmark & \cellcolor{OverallAnomBg}Finance \\
 & \cellcolor{OverallAnomBg}\textbf{FraudDDPM} \cite{pushkarenko2024synthetic} & \cellcolor{OverallAnomBg}2024 & \cellcolor{OverallAnomBg}ISIJ & \cellcolor{OverallAnomBg}Anomaly Detection, Synthesis (single, finance) & \cellcolor{OverallAnomBg}\checkmark & \cellcolor{OverallAnomBg}\checkmark (OH, FT) & \cellcolor{OverallAnomBg}DDPM & \cellcolor{OverallAnomBg}4 & \cellcolor{OverallAnomBg}Utility & \cellcolor{OverallAnomBg}\ding{55} & \cellcolor{OverallAnomBg}\checkmark & \cellcolor{OverallAnomBg}Finance \\
 & \cellcolor{OverallAnomBg}\textbf{DDAE} \cite{sattarov2025diffusion} & \cellcolor{OverallAnomBg}2025 & \cellcolor{OverallAnomBg}KDD & \cellcolor{OverallAnomBg}Anomaly Detection & \cellcolor{OverallAnomBg}\checkmark (ZS) & \cellcolor{OverallAnomBg}\ding{55} & \cellcolor{OverallAnomBg}Denoising Autoencoder & \cellcolor{OverallAnomBg}47/57 A & \cellcolor{OverallAnomBg}PR-AUC, ROC-AUC & \cellcolor{OverallAnomBg}\codeurl{https://github.com/sattarov/AnoDDAE} & \cellcolor{OverallAnomBg}\checkmark & \cellcolor{OverallAnomBg}Generic \\
\cmidrule(lr){1-13}
Flow Matching & \cellcolor{OverallAnomBg}\textbf{TCCM} \cite{li2025scalable} & \cellcolor{OverallAnomBg}2025 & \cellcolor{OverallAnomBg}NeurIPS & \cellcolor{OverallAnomBg}Anomaly Detection & \cellcolor{OverallAnomBg}\checkmark (ZS) & \cellcolor{OverallAnomBg}\ding{55} & \cellcolor{OverallAnomBg}Flow Matching & \cellcolor{OverallAnomBg}47/57 A & \cellcolor{OverallAnomBg}PR-AUC, ROC-AUC, Runtime & \cellcolor{OverallAnomBg}\codeurl{https://github.com/ZhongLIFR/TCCM-NIPS} & \cellcolor{OverallAnomBg}\checkmark & \cellcolor{OverallAnomBg}Generic \\
\bottomrule
\end{tabular}%
}
\end{table*}

\subsubsection{Auxiliary Data Generation for Fraud Detection}
\label{subsubsec:AuxiliaryFraudDetection}

Auxiliary data generation for fraud detection uses diffusion models as supervised oversamplers rather than as direct anomaly scorers. \textbf{FraudDiffuse} \cite{roy2024frauddiffuse} targets labeled transaction data with severe class imbalance. It embeds categorical features into a continuous space and trains a latent DDPM to generate fraudulent transactions, using the normal-transaction distribution to guide the diffusion process toward boundary-like fraud samples and adding a contrastive loss to keep synthetic fraud close to real fraud. This makes diffusion useful for improving downstream fraud classifiers, but it still requires labeled fraudulent data and may miss new fraud patterns outside training.

\textbf{FraudDDPM} \cite{pushkarenko2024synthetic} follows the same supervised augmentation goal but uses a simpler class-conditional strategy: one diffusion model generates normal transactions and another generates fraudulent transactions, after which synthetic samples are merged with real data to build a balanced classifier-training set. This turns diffusion into an oversampling tool for fraud detection. Its key evaluation question is therefore not whether the generated samples look realistic in isolation, but whether they improve downstream detection without leakage, unrealistic transactions, or overfitting to known fraud patterns.

\subsubsection{Comparative Analysis}
\label{subsubsec:AnomalyComparativeAnalysis}
Overall, the methods in Table~\ref{SummaryAnomalyDetection} use diffusion or flow matching in two different ways. The first use is normality scoring. TabADM asks whether a record can be reconstructed from a robust diffusion model even when training data may be contaminated, while SDAD, DDAE, DTE, NSCBAD, and TCCM mainly assume normal-only or mostly normal training data and turn different learned quantities into anomaly scores: latent reconstruction, scheduled denoising, diffusion time, score mismatch, or contraction mismatch. The key design difference is therefore not simply diffusion versus flow matching, but which notion of normality is exposed by the learned dynamics and how strongly that score depends on the chosen feature representation.

The second use is supervised auxiliary generation, where FraudDiffuse and FraudDDPM synthesize minority fraud samples to improve a downstream classifier. Here, the generative model is not itself the detector; it reshapes the training distribution so that the classifier can see more rare-event patterns. The central question is therefore whether synthetic fraud samples add useful variation while preserving transaction semantics, temporal plausibility, and the boundary between normal and fraudulent behavior. A useful direction is to combine the two roles more deliberately: normality scores can provide feature-level evidence at inference time, while controlled synthesis can cover rare patterns that are underrepresented during training.

\section{Discussion and Conclusions}
\label{sec:conclusions}
The methods summarized in Table~\ref{tab:OverallSummary} show that diffusion and flow matching models have begun to address several challenges discussed in Sec.~\ref{Sec:ChallengeswithTabularData}, but the area is still developing. Recent benchmark evidence shows that conclusions about tabular generators can change once feature encoding, hyperparameter search space, tuning budget, sampling cost, and carbon footprint are controlled~\cite{kindji2025tabular}. Broader studies of synthetic tabular data also show that generation quality cannot be separated from post-processing, evaluation, privacy protection, application setting, and deployment constraints~\cite{shi2025comprehensive}. Progress in diffusion and flow matching models therefore depends not only on stronger architectures, but also on clearer data assumptions, reproducible evaluation protocols, and explicit reporting of utility--privacy--cost tradeoffs. We close by summarizing several open directions.

(1) \textbf{Scalability and Sampling Cost}: Diffusion models are typically computationally intensive, especially for high-dimensional or large-scale tabular datasets. Recent work on scalable Forest-Diffusion~\cite{cresswell2024scaling} and flow-matching-based tabular synthesis, including TabSynFlow~\cite{nasution2026flow}, TabFlowM~\cite{anonymous2026tabflowm}, TabCascade~\cite{mueller2026cascaded}, and PatientFlow~\cite{branco2026patientflow}, shows that sampling efficiency should be reported alongside fidelity and utility. Future studies should include training time, sampling time, number of function evaluations, memory footprint, and performance under limited-budget tuning, rather than only final utility or fidelity.

(2) \textbf{Task Coverage Gaps for Flow Matching}: Current tabular flow-matching studies are concentrated in single-table synthesis, longitudinal healthcare synthesis, and anomaly detection. By contrast, multi-relational synthesis, missing-value imputation under incomplete training data, differentially private generation, fairness-aware generation, and memorization mitigation remain largely diffusion-dominated or unexplored from a flow-matching perspective. It remains unclear whether path design, low-cost sampling, and velocity-field residuals can be adapted to missingness mechanisms, relational constraints, formal privacy guarantees, and fairness or memorization objectives.

(3) \textbf{Evaluation, Benchmarking, and Standardization}: Unlike images, tabular data lacks a single perceptual quality metric, so evaluation should jointly cover downstream utility, marginal fidelity, higher-order dependencies, privacy leakage, constraint satisfaction, fairness, and domain validity. Recent benchmark evidence shows that rankings can depend on preprocessing, encoding choices, hyperparameter search space, tuning budget, sampling cost, and carbon footprint~\cite{kindji2025tabular}. Future benchmarks should include missing-data settings, imbalanced classes, high-cardinality categorical features, relational constraints, rare-event domains, and privacy-sensitive settings, while reporting code, preprocessing pipelines, tuning budgets, and failure cases.

(4) \textbf{Privacy, Memorization, and Fairness}: Formal differential privacy has begun to appear in tabular diffusion models~\cite{sattarov2024differentially,sattarov2025privacy}, but most works still rely on empirical privacy metrics such as distance-to-closest-record, attribute inference, or membership inference. Recent memorization studies~\cite{fang2025understanding,fang2025closer} indicate that a small subset of records may dominate memorized generations, so future research should combine formal privacy mechanisms, empirical attack evaluations, and data-centric memorization audits. These analyses should also be connected to fairness, because removing memorization-prone or rare records can change minority-group coverage.

(5) \textbf{Interpretability and Feature-Level Explanation}: Many diffusion and flow matching models operate as black-box generative systems. Anomaly detection methods such as TCCM~\cite{li2025scalable} show that feature-level residuals can support interpretability, but similar explanation tools are underdeveloped for synthesis and imputation. Future methods should expose which features or feature interactions drive generated samples, imputed values, privacy risk, or anomalous scores. These explanations should be connected to tabular preprocessing choices, because scaling, encoding, and inverse transformations can change how feature-level evidence is interpreted.

(6) \textbf{Cross-Modality and Longitudinal Data}: Real data-engineering systems often mix static tables, event sequences, text, images, and time-series measurements. Healthcare models such as FlexGen-EHR~\cite{he2024flexible} and PatientFlow~\cite{branco2026patientflow} point toward multimodal and longitudinal tabular generation, but most current benchmarks remain single-table and static. Extending diffusion and flow matching models to multi-table, temporal, and cross-modal records is therefore an important next step. The key challenge is to preserve alignment between static attributes, temporal events, measurements, and domain-specific validity rules.

(7) \textbf{Hybrid and Representation-Centric Models}: Several recent methods combine diffusion or flow matching with autoencoders, transformers, tree models, or feature-token representations~\cite{suh2023autodiff,zhangmixed,si2026tabrep,zhang2025tabnat,anonymous2026tabflowm}. This trend suggests that the representation layer may be as important as the generative objective. Future work should disentangle representation learning, probability-path design, sampler choice, and backbone effects through controlled ablations. Otherwise, an apparent gain from a new generative model may actually come from better preprocessing, a stronger encoder, or a more favorable inverse mapping.

(8) \textbf{Modeling Feature Correlations and Constraints}: Many methods still handle categorical attributes through one-hot encodings or embeddings and rely on neural networks to recover dependencies implicitly. This is fragile for high-cardinality categories, mixed-type features, relational constraints, and domain rules. Future tabular diffusion and flow matching models should more explicitly model numerical--categorical dependencies, constraint satisfaction, schema-level structure, and rare feature combinations. Otherwise, a generated table may look realistic under marginal statistics while violating valid ranges, business rules, medical logic, or foreign-key consistency.

In summary, diffusion and flow matching models have become promising tools for tabular data modeling, but their advantages depend on representation choices, missingness assumptions, privacy requirements, computational budgets, and evaluation protocols. This survey has reviewed the literature from a task-driven and design-oriented perspective, covering mathematical preliminaries, synthesis and augmentation, imputation, trustworthy data synthesis, anomaly detection, benchmark practice, and open research directions. We hope it helps clarify where the field has made progress and where careful work is still needed.

\bibliographystyle{IEEEtran}
\bibliography{references.bib}

\end{document}